\documentclass[11pt]{article}

\usepackage[preprint]{acl}
\input{config.sty}
\input{glossary.sty}

\title{BioPIE: A Biomedical Protocol Information Extraction Dataset for Experiment Understanding}

\author{
  Haofei Hou$^{*, 1, 3}$, 
  Shunyi Zhao$^{*, 2, 3}$, 
  Fanxu Meng$^{*, 1, 3}$, 
  Kairui Yang$^{1, 3}$, \\
  \textbf{Lecheng Ruan}$^{\textrm{\Letter}, 1, 3}$, 
  \textbf{Qining Wang}$^{\textrm{\Letter}, 1, 3}$
  \vspace{0.5em} \\
  $^{1}$School of Advanced Manufacturing and Robotics, Peking University, Beijing, China\\
  $^{2}$School of Integrated Circuits, Peking University, Beijing, China\\
  $^{3}$PKU-HEC Joint Lab for Industrial Intelligence, Beijing, China\\
  $^*$Equal Contribution\quad\Letter\phantom\,\,\texttt{ruanlecheng@ucla.edu, qiningwang@pku.edu.cn}
}

\begin{document}
\maketitle
\begin{abstract}
Understanding biomedical experiments provides a foundation for downstream tasks, \eg, laboratory automation, and facilitates effective cross-disciplinary communication. Two challenges, \ac{hid} and \ac{msr}, pose unique difficulties for precise automatic experimental understanding. Extracting structured knowledge, \eg, \acp{kg}, is an effective approach to address the \ac{hid} and \ac{msr}. However, existing biomedical datasets for structured knowledge \ac{ie} are limited to a general or coarse-grained level, hindering fine-grained experimental understanding. To address this gap, we introduce \ac{biopie}, a dataset providing procedure-centric \acp{kg} that captures entities, actions, and relations at a scale sufficient for reasoning across biomedical protocols. We evaluate both supervised and \acs{llm}-based \ac{ie} methods on \acs{biopie} to verify its effectiveness, and implement a biomedical question answering system to provide a quantitative illustration of \acs{biopie}'s effectiveness for downstream understanding tasks. The experimental results demonstrate improved understanding performance on both the \ac{hid} and \ac{msr} question sets.
\end{abstract}

\section{Introduction}
Biomedical research spans diverse sub-fields and relies heavily on experimental workflows~\citep{jin2019pubmedqa}. These experiments involve multiple stages, including experiment design, execution, and result analysis. Due to the procedural complexity and wide-ranging domain coverage of biomedical experiments, automated understanding has significant potential for supporting cross-disciplinary collaboration~\citep{rohrbach2022digitization}, laboratory automation, and \ac{ai}-assisted experimental planning~\citep{steiner2019organic, mehr2020universal, burger2020mobile, szymanski2023autonomous, shi2025automated}. However, the language describing these experiments is inherently complex and highly domain-specific, posing major challenges for automation~\citep{frisoni2022text, shi2024autodsl}.

\begin{figure}[t]
\centering
\includegraphics[width=\linewidth]{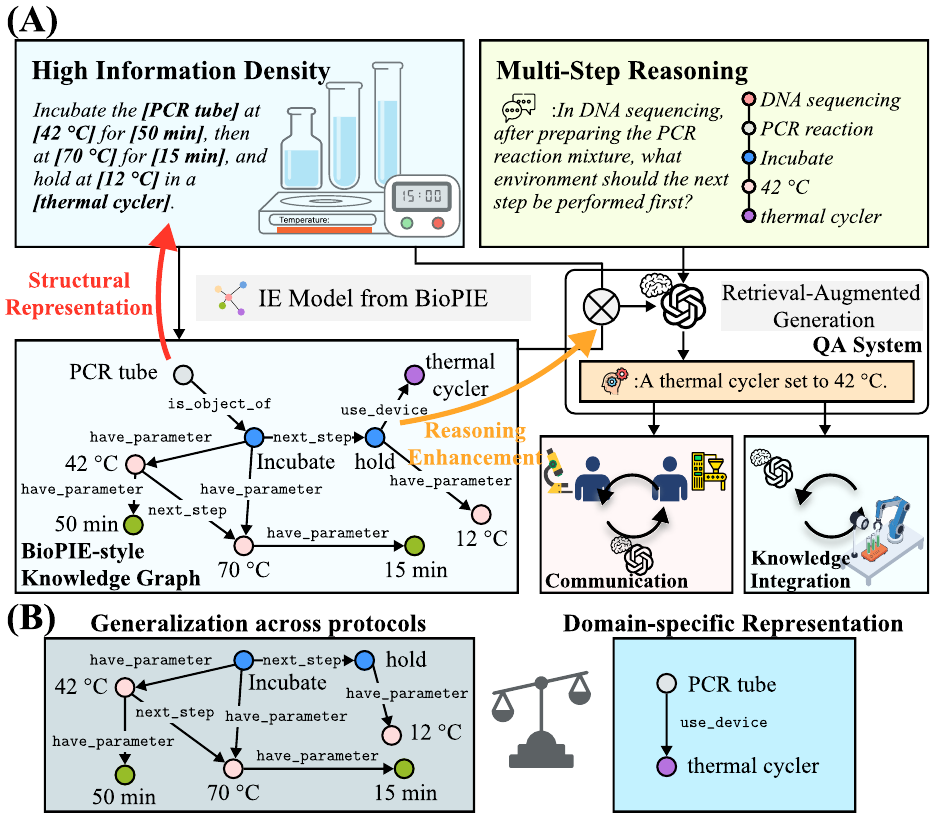}
\caption{\textbf{\ac{biopie} enhances complex biomedical protocol understanding.} \textbf{(A)} The \acsp{kg} in \ac{biopie} provide fine-grained structural representations of experimental steps (\eg, temperature, duration, and execution order), resulting in \textbf{high information density}, and enable \textbf{multi-step reasoning} by integrating sentence-level context with graph-structured knowledge. \textbf{(B)} Existing information extraction datasets involve a trade-off: general datasets lack biomedical knowledge, while domain-specific datasets may not generalize across diverse experiments.}
\label{fig:problem}
\end{figure}

Understanding biomedical experiments exhibits several characteristics. Experimental protocols often condense substantial operational detail into a single sentence~\citep{li2018systematic}, \eg, volumes, temperatures, timings, buffer compositions, and instrument settings, collectively termed \textbf{\acf{hid}}, which requires precise extraction and association with the corresponding operations. Moreover, protocols frequently feature chained steps, hierarchical subroutines, and implicitly distributed information~\citep{ICLR2025_dd9b76f0}, highlighting \textbf{\acf{msr}}: answering a question often necessitates integrating information from multiple distinct steps or facts. Collectively, \ac{hid} and \ac{msr} encompass key aspects of the complexity of biomedical experiment understanding.

Prior work has demonstrated that \acf{ie} of structured knowledge can benefit natural language understanding~\citep{fang-etal-2024-complex}. \acfp{kg} can structurally represent dense and heterogeneous parameters, such as entities, attributes, and their operational relations, thereby making them well-suited to capturing the \ac{hid} and \ac{msr} structures inherent in biomedical experimental protocols. Therefore, a dataset that provides detailed and accurate biomedical experimental \acp{kg} may help natural language understanding of the biomedical protocols. However, there is a lack of datasets specifically designed for this purpose. 

Existing \ac{ie} datasets can be broadly categorized into two groups. Datasets covering general scientific information are broad in sub-fields~\citep{nasar2021named, zhao2024comprehensive}. For example, SciERC and SciER annotate entities such as Method, Task, Metric, and Material, along with relations including \texttt{Used-for}, \texttt{Part-of}, \texttt{Compare}, and \texttt{Evaluate}~\citep{luan-etal-2018-multi, zhang2024scier}. However, the lack of domain-specific text makes it difficult for these datasets to fully represent the experimental reagents, materials, containers, and devices required in biomedical applications, as shown in Fig.~\ref{fig:problem}\textbf{(B)}.

The other category comprises datasets specifically designed for biomedical research~\citep{arsenyan2024large, peng2024depth}. These biomedical datasets primarily focus on entities such as proteins, chemicals, drugs, and diseases~\citep{kringelum2016chemprot, krallinger2021drugprot}, with relation types covering the molecular and pharmacological interactions~\citep{HERREROZAZO2013914, zhang2022distant, luo2022biored, lai2024enzchemred}. Nevertheless, existing biomedical datasets typically provide a relatively coarse-grained scheme centered on molecular and pharmacological entities, lacking the procedural granularity required for biomedical experiment understanding. Together, these observations highlight the need for a dataset that encompasses diverse, cross-disciplinary experimental protocols with sufficient procedural detail to support biomedical experiment understanding~\citep{perera2020named}. To the best of our knowledge, such a dataset does not currently exist.

In this paper, we introduce \acf{biopie}, a new dataset specifically designed to support biomedical experiment understanding. \ac{biopie} provides clearly defined biomedical experimental protocols and corresponding \acp{kg}, with broad cross-disciplinary coverage of experimental entities, actions, and procedural relations. It is constructed to support generalizable reasoning and machine understanding of biomedical protocols, thereby improving biomedical experimental understanding, which is illustrated in Fig. \ref{fig:problem}\textbf{(A)}.

The contributions of this paper are as follows: (1) We construct \ac{biopie}, an \ac{ie} dataset for understanding complex biomedical experiments, including both \ac{hid} and \ac{msr} aspects; (2) We systematically evaluate different \ac{ie} algorithms on \ac{biopie}, including both supervised models and \acp{llm} under different settings; and (3) We develop a \acf{qa} system to provide a quantitative illustration that \ac{biopie} effectively supports biomedical experiment understanding, particularly for \ac{hid} and \ac{msr}.

\section{\acs{biopie} Dataset}

\subsection{Data Scheme}

\begin{figure*}[t]
\centering
\includegraphics[width=\textwidth]{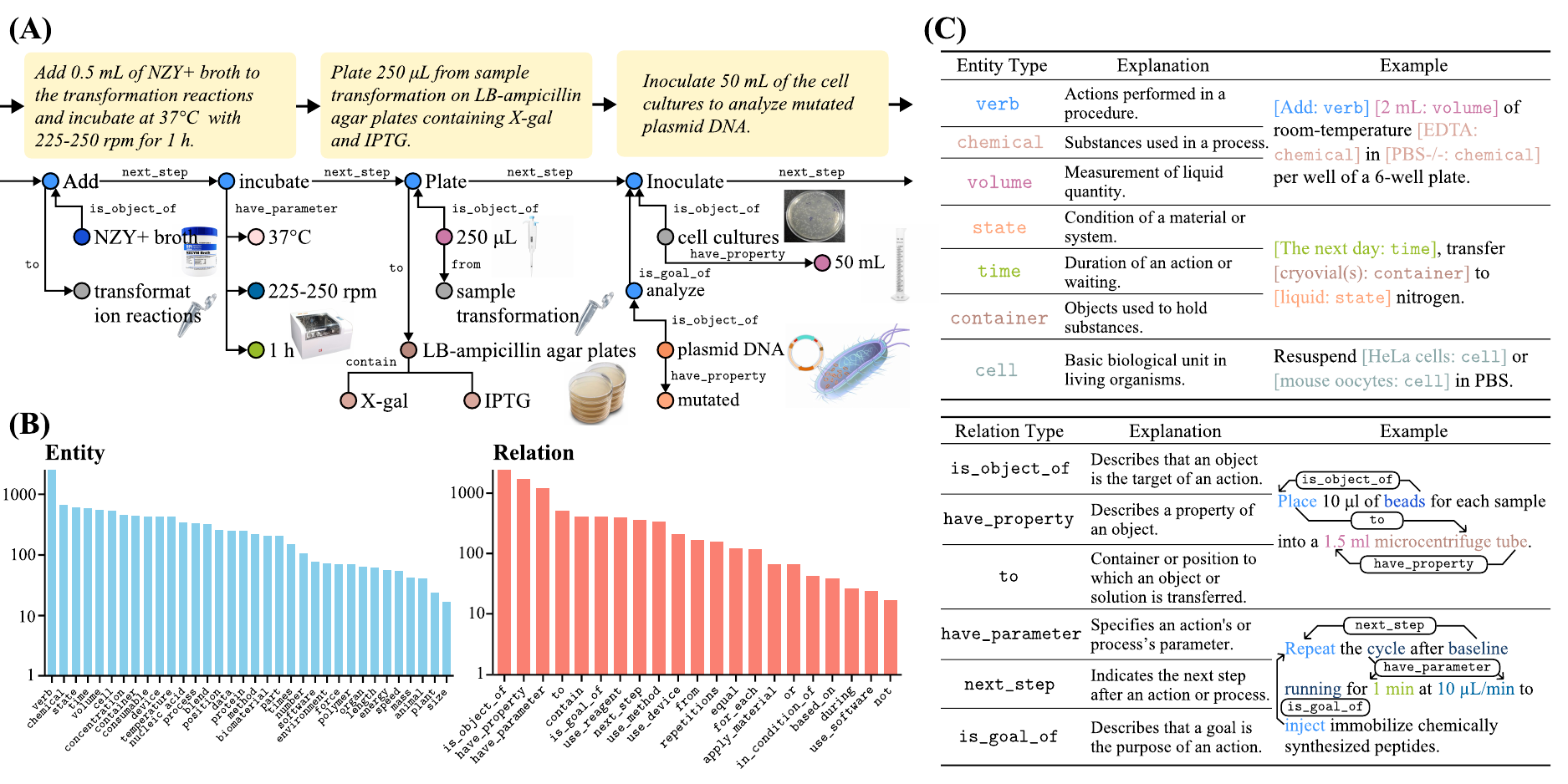}
\caption{\textbf{Overview of \acs{biopie}.} \textbf{(A)} An annotated plasmid DNA preparation protocol illustrating how domain-specific experimental descriptions are mapped into a shared, procedure-centric representation of actions, materials, parameters, and procedural dependencies, thereby supporting consistent annotation across diverse biomedical workflows. \textbf{(B)} Distributions of the 34 entity types and 21 relation types in the dataset. \textbf{(C)} Definitions and examples of representative entity and relation labels, with examples drawn from different protocols to demonstrate their applicability across biomedical procedures.}
\label{fig:dataset}
\end{figure*}

To support robust knowledge extraction from biomedical protocols, we design an annotation scheme that captures the essential operational structure while avoiding overly domain-specific categories. The scheme is fine-grained for procedural reasoning yet generalizable across biomedical experimental workflows rather than being tied to a narrow subdomain, as shown in Fig.~\ref{fig:dataset}\textbf{(A)}, which is different from Fig.~\ref{fig:problem}\textbf{(B)}. The detailed design and development process of ontologies can be found in Appx.~\ref{apxsec:ontologies}.

\ac{biopie} comprises 34 entity types encompassing actions, materials, laboratory instruments, biological samples, and key experimental parameters (\eg, \texttt{time}, \texttt{temperature}, and \texttt{force}). \ac{biopie} not only provides general definitions for actions and processes, but also focuses on operational elements that recur across diverse biomedical procedures, \eg, centrifugation forces, incubation temperatures, buffers, and consumables. Following standard practice in scientific \ac{ie}~\citep{stenetorp2012brat, zhang2024scier}, annotators allow nested spans when necessary for relation attachment.

We define 21 relation types to describe how experimental entities interact within a protocol, including action–object relations (\texttt{is\_object\_of}), action–parameter relations (\texttt{have\_parameter}), resource-usage relations (\texttt{use\_device}), structural relations (\texttt{contain}), and procedural logic (\texttt{next\_step}). 
These relations capture the diverse aspects of human instructions in experimental protocols without relying on domain-specific biological semantics, making them suitable for heterogeneous protocols. Some detailed explanations of representative entity and relation labels are provided in Fig.~\ref{fig:dataset}\textbf{(C)}.

Our balanced design allows the \acp{kg} to support understanding while maintaining broad applicability across diverse biomedical experiments. First, our scheme is procedure-centric and focuses on operational details such as actions, materials, and parameters. At the same time, it deliberately avoids narrowly specialized biomedical categories, enabling consistent annotation across cell culture protocols, microscopy workflows, sequencing preparations, biomaterial fabrication, and other experimental contexts. \ac{biopie} provides a comprehensive representation of biomedical workflows, achieved through these detailed, fine-grained local representations. Detailed examples can be found in Appx. \ref{apxsec:comprehensive}. 

\subsection{Data Collection and Processing}

We first collect protocols from peer-reviewed biology research articles and use Qwen-Max to clean and normalize them into stepwise imperative sentences. From these protocols, we select 344 sub-protocols covering four major domains: genetic manipulation, molecular interactions, physiological signal detection, and regenerative medicine, from the complete collection as our \ac{id} set, covering a broad range of common biomedical workflows. To construct an \ac{ood} set, we select the other 45 sub-protocols whose biomedical sub-fields are not represented in the \ac{id} set. These \ac{ood} protocols cover distinct experiment types such as plant-based biopharmaceuticals, functional imaging, and epidemiological analysis. We also analyze the detailed distribution of data sources (Appx.~\ref{apxsec:datasource}), and the impact of this normalization of protocols qualitatively and quantitatively (Appx.~\ref{apxsec:normalization}). The analyses demonstrate the normalization has no significant impact on the information within experimental workflows. 
\subsection{Data Annotation}

Two annotators with graduate-level backgrounds in computer science and biomedical research are recruited. The annotators receive training before starting the task. The dataset was annotated through an independent parallel annotation process by two annotators who had no access to each other's results during the task. A lead annotator with the same background then reconciled all discrepancies according to annotation guideline to produce the final gold-standard labels. For all protocols, we compute inter-annotator agreement using Cohen’s kappa~\citep{davies1982measuring}. The kappa score for entity annotation is 79.20\% and for relation annotation is 68.26\%, achieving a level of consistency comparable to that reported in existing literature~\citep{luan-etal-2018-multi, zhang2024scier}.

\subsection{Data Statistics}

\begin{table}[t]
\centering
\resizebox{\linewidth}{!}{%
\begin{tabular}{lccccc}
\toprule
& ACE2005 & SciERC & ChemProt & \ac{biopie} \\
\midrule
Entity Types & 7 & 6 & 3 & 34 \\
Relation Types & 6 & 7 & 11 & 21 \\
Entities & 38287 & 8089 & 17340 & 10982 \\
Relations & 7070 & 4716 & 10065 & 8848 \\
Sentences & 10372 & 2679 & 7552 & 1916 \\
Relations/Sent. & 0.68 & 1.76 & 1.33 & 4.62 \\
\bottomrule
\end{tabular}
}
\caption{\textbf{Comparison of \ac{biopie} and three datasets. %supporting \ac{ie} in scientific text or biomedical literature.
}}
\label{tab:dataset}
\end{table}

As shown in Tab. \ref{tab:dataset}, \ac{biopie} contains over 10.9k entities and 8.8k relations, which are comparable in scale to existing \ac{ie} datasets~\citep{Doddington2004TheAC, kringelum2016chemprot, luan-etal-2018-multi}. \ac{biopie} exhibits substantially higher relational density than prior datasets, averaging 4.6 relations per sentence compared to 0.7~\citep{Doddington2004TheAC}–1.7~\citep{luan-etal-2018-multi} in existing corpora. This reflects the inherently structured and interaction-rich nature of biomedical protocols. High relational density directly indicates information density but does not necessarily imply multi-step logical chain complexity. We randomly split the dataset into the training, development, and \ac{id} test sets using a 7:1:2 ratio. The additional protocols are used as the \ac{ood} test set. \ac{id} evaluation assesses model reasoning on seen objects and protocols under similar distributions. Conversely, \ac{ood} protocols cover entirely unseen biomedical fields and experiment types, serving as a true bottleneck to test cross-domain generalization. Fig. \ref{fig:dataset}\textbf{(B)} presents the detailed distribution of each entity and relation type.

\section{\acs{biopie} Benchmarking}

\subsection{\acl{ie} Baselines}\label{subsec:baselines}

We consider two commonly used types of \ac{ie} methods: supervised models, which exhibit strong \ac{ie} performance on specific tasks; and \acp{llm}, which are pretrained on broad-coverage corpora and provide more general \ac{ie} capabilities~\citep{chang2024survey, naveed2025comprehensive}. For both supervised models and \acp{llm}, we investigate two architectural frameworks: a pipeline framework, which performs \ac{ner} and \ac{re} separately, and a joint \ac{ere} framework, which performs \ac{ner} and \ac{re} jointly.

With the above baseline selection criteria, we select two \ac{sota} supervised models as baselines: PL-Marker~\citep{ye2022packed}, which adopts a span-based representation strategy within a pipeline framework, and HGERE~\citep{yan2023joint}, which introduces a joint \ac{ere} framework. Considering the zero-shot, few-shot, and \ac{lora}~\citep{hu2022lora} settings of \acp{llm}, we combine each setting with the pipeline and joint frameworks, resulting in six \ac{llm}-based configurations in total. Under both zero-shot and few-shot settings, we evaluate GPT-5, Claude-4.5-Opus, Llama-4, and Qwen-max. Under the \ac{lora} setting, we evaluate Llama-3-8B and Qwen-3-8B. 

\subsection{\acl{ie} Evaluation Details}\label{subsec:implementation_settings}

For supervised baselines, we use \textit{scibert}~\citep{beltagy2019scibert} and \textit{biobert}~\citep{lee2020biobert} as the encoder. In the few-shot setting for \acp{llm}, we employ a sentence retriever to select the most similar training examples as in-context demonstrations~\citep{dong2024survey}. For each task, we retrieve up to 20 candidate demonstrations and select the number of demonstrations that yield the highest Rel+ score on the validation set. We use OpenAI's \textit{text-embedding-3-large} model as the retriever backbone. The instruction part of our prompt is adapted from Chat\ac{ie}~\citep{wei2023chatie}, and we additionally provide component label definitions to improve clarity and model understanding~\citep{zhang2024scier}. The complete prompt can be found in Appx. \ref{apxsec:guideline}. During our experiments, the random seed is set to zero.

Given an input protocol $D$ with sentences $\mathcal{S} = \{s_1, s_2, \ldots, s_N\}$, we define \ac{ie} independently at the sentence level as follows. Let $\mathbb{E}$ denote a set of entity types. Given a sentence $s_i = \{w_1, w_2, \ldots, w_k\}$, the \ac{ner} task identifies a set of entity mentions $\{e_1, e_2, \ldots, e_m\}$.
Each entity mentioned $e_j = \{w_l, \ldots, w_r\}$ corresponds to a contiguous span of tokens and is assigned an entity type $t_j \in \mathbb{E}$. Let $\mathbb{R}$ denotes the set of relation types. The \ac{re} task predicts a relation label $r_{jk} \in \mathbb{R} \cup \{\texttt{NULL}\}$ for each ordered entity pair $(e_j, e_k)$ occurring within sentence $s_i$. The special label \texttt{NULL} indicates the absence of a semantic relation. In our \ac{re} task, we require \acp{llm} to explicitly generate \texttt{NULL} to mitigate their tendency to misclassify non-relational pairs into predefined categories~\citep{zhang2024scier}. However, to ensure rigorous evaluation, these \texttt{NULL} predictions are strictly excluded from the final metric calculations.

\begin{table*}[t]
\centering
\resizebox{0.80\textwidth}{!}{%
\begin{tabular}{lcccccccc}
\toprule
\multirow{2}{*}{} &  \multicolumn{4}{c}{In-domain} & \multicolumn{4}{c}{Out-of-domain} \\
& \ac{ner} & Rel & Rel+ & \ac{re} & \ac{ner} & Rel & Rel+ & \ac{re} \\
\midrule
\multicolumn{9}{c}{\textit{Supervised Baselines}} \\
\midrule
PL-Marker$^1$~\citep{ye2022packed} & 87.40 & \textbf{82.55} & \textbf{74.52} & \textbf{87.88} & 73.87 & 70.27 & 52.27 & \textbf{78.85} \\
PL-Marker$^2$~\citep{ye2022packed} & 86.33 & 80.45 & 72.48 & 33.69 & 51.93 & 23.51 & 16.52 & 24.79 \\
HGERE$^1$~\citep{yan2023joint}     & \textbf{87.63} & 82.10 & 73.93 &  -  & 74.58 & \textbf{70.49} & \textbf{52.41} &  -  \\
HGERE$^2$~\citep{yan2023joint}     & 86.95 & 80.93 & 73.08 & - & 72.76 & 69.48 & 51.48 & - \\
\midrule
\multicolumn{9}{c}{\textit{Zero-shot \ac{llm}}} \\
\midrule
GPT-5 (Pipeline)         & 57.14 & 50.47 & 41.14 & 69.86 & 52.08 & 51.46 & 37.60 & 68.24 \\
GPT-5 (Joint)            & 22.90 & 22.23 & 17.54 &   -   & 21.94 & 21.23 & 14.78 &   -   \\
Claude-4.5-Opus (Pipeline)    & 69.34 & 40.56 & 31.90 & 48.01 & 63.81 & 34.37 & 24.90 & 43.33 \\
Claude-4.5-Opus (Joint)       & 39.37 & 33.66 & 26.11 &   -   & 65.74 & 31.31 & 24.79 &   -   \\
Llama-4 (Pipeline)       & 41.08 & 1.88  & 1.44  & 1.69  & 42.34 & 1.56  & 1.21  & 0.74  \\
Llama-4 (Joint)          & 1.73  & 0.75  & 0.00  &   -   & 0.92  & 0.38  & 0.19  &   -   \\
Qwen-max (Pipeline)      & 67.12 & 20.53 & 17.02 & 27.38 & 60.23 & 8.11  & 13.05 & 25.05 \\
Qwen-max (Joint)         & 65.73 & 24.20 & 19.10 &   -   & 61.34 & 20.02 & 14.05 &   -   \\
\midrule
\multicolumn{9}{c}{\textit{Few-shot \ac{llm}}} \\
\midrule
GPT-5 (Pipeline)         & 62.74 & 66.83 & 59.73 & 79.30 & 52.80 & 54.00 & 41.42 & 68.46 \\
GPT-5 (Joint)            & 27.71 & 25.03 & 22.14 &   -   & 33.89 & 32.18 & 26.08 &   -   \\
Claude-4.5-Opus (Pipeline)    & 85.18 & 67.87 & 63.47 & 77.20 & 73.23 & 53.38 & 41.48 & 64.88 \\
Claude-4.5-Opus (Joint)       & 83.41 & 65.88 & 60.27 &   -   & 73.11 & 52.57 & 41.33 &   -   \\
Llama-4 (Pipeline)       & 49.05 & 18.73 & 16.75 & 21.03 & 41.23 & 11.71 & 9.73  & 16.24 \\
Llama-4 (Joint)          & 13.51 & 7.73  & 7.39  &   -   & 18.26 & 10.15 & 7.69  &   -   \\
Qwen-max (Pipeline)      & 83.28 & 64.35 & 59.87 & 73.17 & 71.81 & 46.23 & 36.45 & 56.74 \\
Qwen-max (Joint)         & 75.68 & 55.49 & 51.84 &   -   & 62.16 & 37.25 & 29.27 &   -   \\
\midrule
\multicolumn{9}{c}{\textit{\acs{lora} \ac{llm}}} \\
\midrule
Llama-3-8B (Pipeline)    & 86.33 & 75.28 & 68.13 & 81.67 & \textbf{75.70} & 62.97 & 49.44 & 73.24 \\
Llama-3-8B (Joint)       & 84.71 & 74.86 & 66.58 &   -   & 74.86 & 62.68 & 46.72 &   -   \\
Qwen-3-8B (Pipeline)     & 84.44 & 69.95 & 62.68 & 77.06 & 74.90 & 61.63 & 46.81 & 68.59 \\
Qwen-3-8B (Joint)        & 82.92 & 70.70 & 62.96 &   -   & 73.86 & 59.99 & 44.54 &   -   \\
\bottomrule
\end{tabular}
}
\caption{\textbf{Test F1 scores of different baselines on our proposed dataset.} ``Joint'' denotes joint \acs{ie}, while ``Pipeline'' refers to performing \acs{ner} and \acs{re} separately. ``Rel'' and ``Rel+'' indicate relation extraction from original text under boundary and strict evaluation, respectively, and ``\ac{re}'' denotes relation extraction with gold entities, applicable only to the pipeline methods. The number of in-context examples was tuned on the validation set, as shown in Fig. \ref{fig:iecompare}\textbf{(A)}. $^1$ Denotes the use of \textit{scibert}~\citep{beltagy2019scibert} as the encoder; $^2$ denotes \textit{biobert}~\citep{lee2020biobert}. } 
\label{tab:iebenchmark}
\end{table*}

The evaluation tasks include \ac{ner}, \ac{re} from original text, and \ac{re} with gold standard entities. For \ac{ner}, we conduct span-level evaluation, requiring both correct boundaries and entity types. For \ac{re} from original text, we report two metrics following prior work~\citep{ye2022packed, yan2023joint}: (1) Boundary evaluation (Rel), which requires correct prediction of subject and object boundaries and their relation, and (2) Strict evaluation (Rel+), which additionally requires correct entity types. 

\subsection{\acl{ie} Results}\label{subsec:biopie_results}

\begin{figure}[t]
\centering
\includegraphics[width=\linewidth]{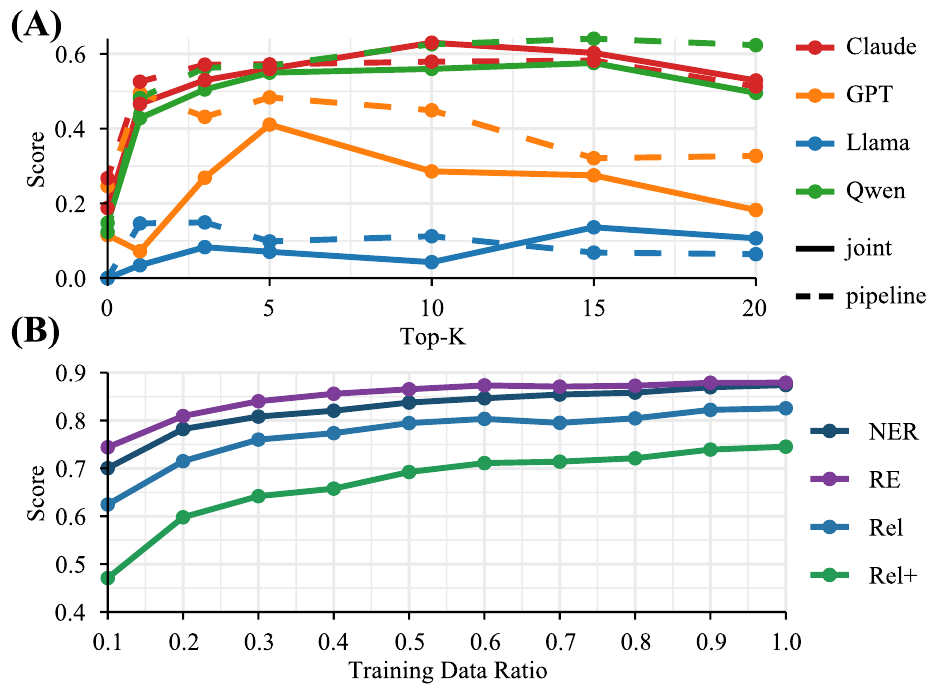}
\caption{\textbf{Effects of settings on \ac{ie} methods.} \textbf{(A)} Impact of the number of the retrievals on the validation set for Rel+ F1 score. \textbf{(B)} Performance trends of PL-Marker under varying training-protocol proportions on the \ac{id} test set.}
\label{fig:iecompare}
\end{figure}

Tab.~\ref{tab:iebenchmark} reports the experimental results on both the \ac{id} and \ac{ood} test sets. Fig.~\ref{fig:iecompare}\textbf{(A)} illustrates the impact of varying the number of demonstrations in the few-shot setting on \ac{re} from the original text performance (Rel+). Overall, introducing a small number of demonstrations yields substantial performance gains. Most \acp{llm} reach their peak performance with approximately 5–15 demonstrations, after which additional examples provide diminishing or even negative returns. These findings suggest that overly large demonstration sets may introduce noise and reduce the effectiveness of \ac{icl}. Detailed results for precision and recall are provided in the Appx. \ref{apxsec:detailie}.

Among supervised baselines using \textit{scibert}, PL-Marker achieves the best \ac{id} performance (82.55 Rel, 74.52 Rel+, 87.88 \ac{re}). In contrast, HGERE demonstrates stronger \ac{ood} robustness (74.58 \ac{ner}, 70.49 Rel, 52.41 Rel+). For both models, \ac{id}-to-\ac{ood} performance drops more for \ac{ner} ($\sim$13) than for \ac{re} (10–12), indicating unseen entities are harder to recognize. Notably, substituting \textit{scibert} with \textit{biobert} causes a catastrophic \ac{ood} collapse in PL-Marker (\eg, Rel drops to 23.51), whereas HGERE remains robust (69.48 Rel). This suggests that HGERE's \ac{ood} advantage is an intrinsic architectural benefit rather than solely a byproduct of the pretrained encoder.

In the zero-shot setting, \acp{llm} exhibit substantial performance variability. GPT-5 achieves the most balanced pipeline performance on \ac{id} data (57.14 \ac{ner}, 41.14 Rel+), while Claude-4.5-Opus achieves strong \ac{ner} performance but weaker \ac{re} results. Llama-4 performs poorly across most \ac{re}-related metrics, and Qwen-max achieves reasonable \ac{ner} performance but limited \ac{re} capability. Across all models, pipeline extraction outperforms joint extraction, highlighting the benefit of decomposing \ac{ner} and \ac{re} for \acp{llm}. Compared to supervised baselines, \acp{llm} show more consistent performance between \ac{id} and \ac{ood}, likely due to their large-scale pretraining.

Few-shot learning yields dramatic improvements across all evaluated models. Claude-4.5-Opus with pipeline extraction achieves the largest gains, reaching 85.18 (\ac{ner}) and 63.47 (Rel+) on \ac{id} data. However, improvements from \ac{icl} are generally larger on \ac{id} than \ac{ood}, as demonstrations are more similar to \ac{id} samples.

\ac{lora}-tuned smaller \acp{llm} demonstrate that parameter-efficient fine-tuning can rival or even surpass \acp{llm} with \ac{icl}. Llama-3-8B with pipeline extraction achieves 86.33 (\ac{ner}) and 68.13 (Rel+) on \ac{id}, approaching supervised performance while exhibiting strong generalization. Although pipeline extraction remains slightly superior to joint extraction after fine-tuning, the gap becomes much smaller.

Fig.~\ref{fig:iecompare}\textbf{(B)} shows the performance trends of PL-Marker on \ac{ner}, Rel, Rel+, and \ac{re}, with the different training scales. As the dataset size increases, \ac{re} performance improves more slowly and gradually saturates, while \ac{ner} continues to show moderate gains. Rel and Rel+ also consistently improve with more training data. In particular, Rel+ increases from 47.09 (0.1 of the training set) to 59.79 (0.2), 69.24 (0.5), 73.92 (0.9), and 74.52 (1.0), with diminishing gains as the data scale grows.

\section{Biomedical Experiment Understanding Validation}

As a typical natural language understanding task, biomedical experiment \acf{qa} has become an active research area in recent years~\citep{liu2025bioprobench}. To demonstrate the utility of \ac{biopie} in enhancing experimental understanding, we implement a \ac{kg}-augmented \ac{qa} system and evaluate its performance on questions specifically designed to test \ac{hid} and \ac{msr} capabilities. 

\subsection{\acs{qa} System Design}

The proposed \ac{qa} system follows current basic paradigms for graph-based reasoning in \acp{llm}~\citep{sun2023think}. We adopt a synergistic architecture that jointly leverages unstructured text and structured graph knowledge to enhance response fidelity. To ensure computational efficiency and leverage the high density of specialized terminology in biomedical protocols, we first implement a filter. This module evaluates candidate sentence–graph pairs $(s_i, g_i)$ for a query $q$ using a hybrid relevance score:
\begin{equation}
R(q, s_i, g_i) = R_t(q, s_i) \cdot \log(1 + R_g(q, g_i)),
\end{equation}
where $R_t(q, s_i)$ denotes the semantic similarity between the query and the sentence, and $R_g(q, g_i) = \sum_{v \in \mathcal{V}_i} \mathbb{I}[v \subseteq q]$ quantifies the structural alignment by measuring the entity overlap between the query and the knowledge graph $g_i$ ($\mathcal{V}_i$ denotes the entity set of $g_i$) ~\citep{kruit2024retrieval}.

Building upon this filtered context, the planner module decomposes complex natural language queries into structured execution paths. Specifically, it models the generation of an abstract relation sequence $z = \{r_1, \dots, r_l\}$ via an \ac{llm} as $P_\theta(z \mid q, \hat{s}, \hat{G})$, where $\hat{s}$ and $\hat{G}$ denote the sets of retrieved sentences and graphs, respectively. Guided by these plans, the retriever serves as a grounding engine, instantiating the abstract paths within the knowledge graph to identify specific trajectories $w_z = e_q \xrightarrow{r_1} e_1 \dots \xrightarrow{r_l} e_a$ that link query entities to the potential answers.

\begin{figure}[t]
\centering
\includegraphics[width=\linewidth]{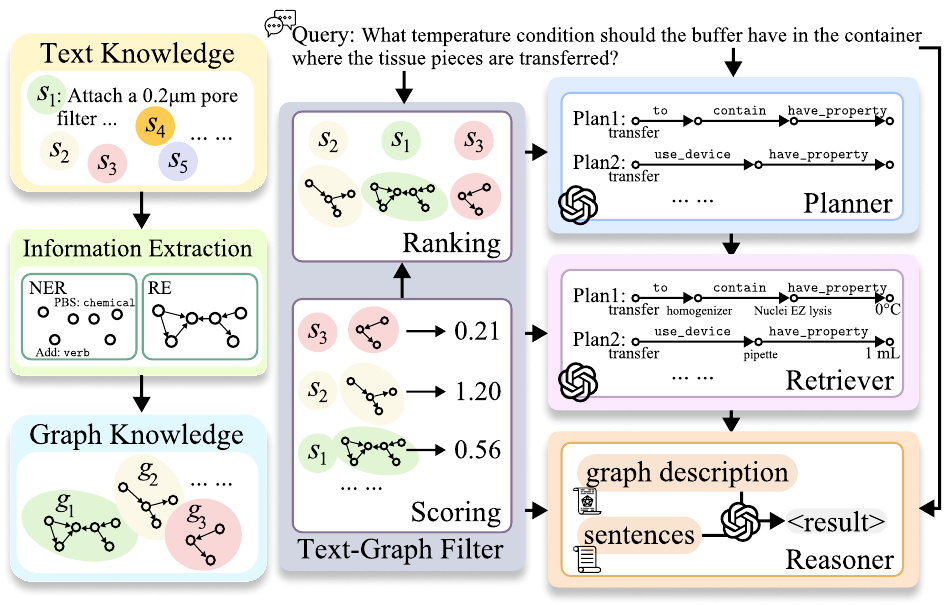}
\caption{\textbf{Pipeline of the proposed \ac{qa} system.} }
\label{fig:rag}
\end{figure}

In the final stage, the reasoner synthesizes the retrieved textual evidence $\hat{s}$ with the instantiated graph descriptions $G_z=\{w_z\}$ to formulate the terminal response $Y$. The generation process, parameterized by $\theta$, is optimized by maximizing the conditional probability:
\begin{equation}
p_\theta(Y \mid q, \hat{s}, G_z) = \prod_{t=1}^{|Y|} P_\theta(y_t \mid y_{<t}, [q, \hat{s}, G_z]).
\end{equation}By integrating structured logical constraints with rich linguistic context, the system ensures that the output is consistent with both textual and relational data, as illustrated in Fig. \ref{fig:rag}.

\subsection{\acs{qa} Baselines}\label{subsec:ragbaselines}
To demonstrate the effectiveness of the proposed dataset and \ac{qa} system, we compare our approach against a broad range of commonly used retrieval-based \ac{qa} systems. Specifically, our experiments cover text-based \ac{qa} systems equipped with different retrievers, including BM25~\citep{robertson2009probabilistic}, LaBSE~\citep{feng2022language}, OpenAI's \textit{text-embedding-3-large} (Emb-3-large), and Qwen's \textit{embedding-v4} (Emb-v4). We also include GRAG~\citep{hu-etal-2025-grag}, which relies solely on knowledge-graph-based retrieval without using raw textual corpora, for comparison, and ToG~\citep{sun2023think}, which allows \acp{llm} to iteratively explore and prune reasoning chains.

Furthermore, to investigate the impact of \acp{kg} on biomedical experiment \ac{qa}, we compare graphs constructed from \ac{ie} models trained on different datasets, including SciERC~\citep{luan-etal-2018-multi}, which focuses on scientific \ac{ie},
and ChemProt~\citep{kringelum2016chemprot}, which targets chemical reaction \ac{ie}. In addition, we include two \ac{llm}-only baselines that do not leverage any externally retrieved knowledge for \ac{llm}: a frozen \ac{llm}, and an \ac{llm} fine-tuned using \ac{lora} \citep{hu2022lora}.

\subsection{\acs{qa} Evaluation Details}\label{subsec:ragimplement}

\begin{table}[t]
\centering
\resizebox{0.80\linewidth}{!}{%
\begin{tabular}{lccc}
\toprule
& Test & \ac{hid} & \ac{msr}\\
\midrule
\ac{llm} only & 14.74 & 16.09 & 15.45 \\
\ac{llm} \acs{lora} & 12.44 & 11.49 & 9.76 \\
\midrule
BM25        & 63.72 & 65.52 & 52.03 \\
LaBSE       & 55.60 & 63.22 & 53.66\\
Emb-3-large & 61.52 & 66.09 & 56.10 \\
Emb-v4      & 59.14 & 67.24 & 54.47 \\
\midrule
GRAG                    & 8.83 & 10.34 & 7.32 \\
GRAG \acs{lora}         & 22.77 & 17.82 & 16.26 \\
ToG               & 69.90 & 75.29 & 69.92 \\
\midrule
Ours w/o Sentence       & 59.84 & 64.94 & 65.04 \\
Ours w/o Graph          & 55.69 & 66.67 & 60.16 \\
Ours w/o Planner  & 72.29 & 73.56 & 62.60 \\
Ours w SciERC           & 62.40 & 70.11 & 59.35 \\
Ours w ChemProt   & 64.70 & 72.99 & 64.23 \\
Ours                    & \textbf{72.99} & \textbf{83.33} & \textbf{74.80} \\
\bottomrule
\end{tabular}
}
\caption{\textbf{Performance comparison across different \ac{qa} systems.}}
\label{tab:ragforcomplex}
\end{table}

The \ac{qa} dataset is divided into training, validation, and test sets with sizes of 1983, 159, and 1133, respectively. To further analyze model performance under challenging conditions, we construct \ac{msr} and \ac{hid} question sets from the test set. The detailed \ac{qa} dataset information can be found in Appx. \ref{apxsec:qadatasetdetail}. In the evaluation phase, adhering to the standard paradigm of \ac{kgqa}, relevant contexts are retrieved from the comprehensive pool of all protocols~\citep{luo2023reasoning, sun2023think}. 

We adopt accuracy as the evaluation metric for all experiments, \ie, whether the answer appears in the model output. This is a widely recognized standard~\citep{krithara2023bioasq}. Following the established evaluation paradigm in generative \ac{qa} benchmarks~\citep{wang2023evaluating}, accuracy is initially defined via substring matching to prevent severe false negative penalties caused by verbose or explanatory \ac{llm} outputs~\citep{adlakha2024evaluating}. Furthermore, we tune the number of in-context examples on the validation set. As an additional metric, we compute the retrieval hit rate on the validation set. 

During the experiments, the random seed is set to zero. The sentence-level relevance function $R_{t}(q, s_i)$ is BM25~\citep{robertson2009probabilistic} in implementation. Experiments are conducted on Llama-3-8B. The prompts used for the planner, the retriever, and the reasoner are adapted from RoG~\citep{luo2023reasoning}. Specifically, we use HGERE~\citep{yan2023joint} as the \ac{ie} method, including \ac{ner} and \ac{re}. Furthermore, we conduct ablation studies comparing textual inputs and knowledge-graph-based inputs, while keeping the retrieval strategy fixed to the proposed pipeline.

\subsection{\acs{qa} Evaluation Results}\label{subsec:ragresults}

\begin{figure}[t]
\centering
\includegraphics[width=\linewidth]{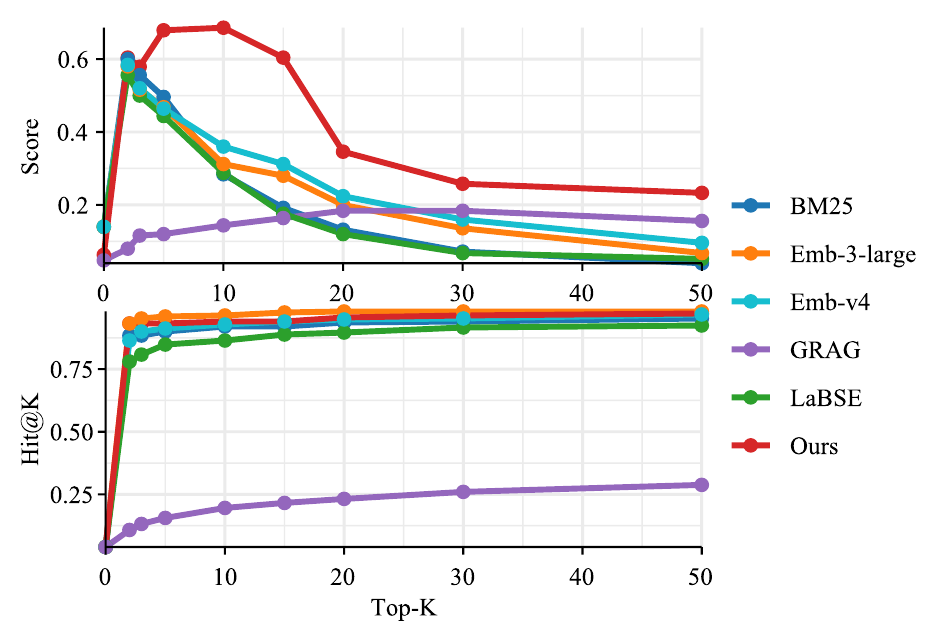}
\caption{\textbf{Effect of the number of retrievals on \ac{qa} systems performance on the validation set.}}
\label{fig:qacompare}
\end{figure}

\begin{figure}[t]
\centering
\includegraphics[width=\linewidth]{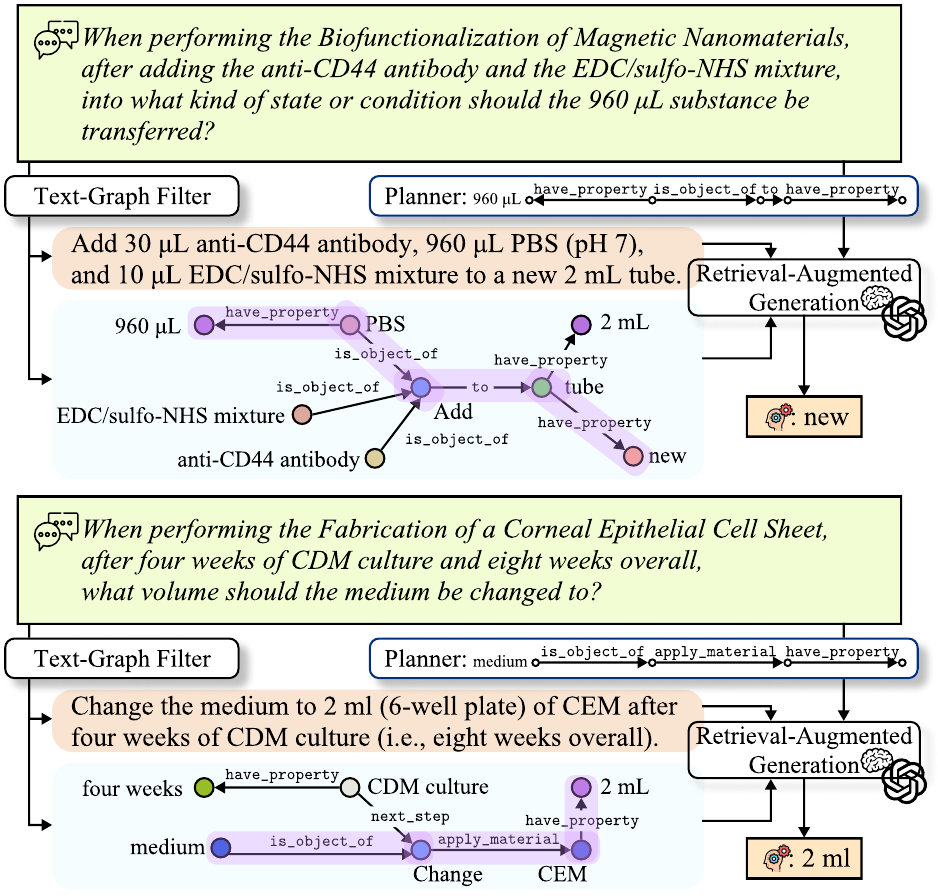}
\caption{\textbf{\ac{qa} system showcases.} }
\label{fig:qashowcases}
\end{figure}

Tab.~\ref{tab:ragforcomplex} presents results on the test set. Fig. \ref{fig:qacompare} shows the effect of varying the number of retrieved texts for accuracy and hit rate on the validation set. Across all settings, the proposed \ac{qa} system achieves the best overall performance. Fig.~\ref{fig:qashowcases} illustrates two example outputs from our experimental \ac{qa} system; additional examples are provided in the Appx. \ref{apxsec:qashowcases}. Detailed evaluation results, including the impact of different hops and various \ac{ie} methods, are provided in Appx. \ref{apxsec:detailqa}. 

Ablation studies further confirm the complementary role of graph-based inputs. Removing graph-based knowledge (``Ours w/o Graph'') leads to a noticeable performance degradation compared to the full model. Our findings demonstrate that combining sentences with structured \acp{kg} can effectively address the challenges of \ac{hid} and \ac{msr} to achieve optimal performance. To isolate specific component contributions, we hold the \ac{kg} fixed and observe that advanced graph reasoners like ToG (69.90\%) outperform standard retrievers like BM25 (63.72\%); further, our full reasoning planner (72.99\%) outperforms the planner ablation (72.29\%, with a severe \acs{msr} drop to 62.60\%).

Using graphs constructed from SciERC and ChemProt also results in reduced performance, only marginally outperforming the text-retrieval baseline. Specifically, under the identical \ac{qa} pipeline, the \ac{biopie} schema (72.99\% on Test) outperforms SciERC (62.40\%) and ChemProt (64.70\%). This indicates that fine-grained knowledge representations tailored to biomedical experimental protocols are critical for effective biomedical experiment \ac{qa}. As our evaluation protocol targets on unseen text, a Gold \ac{kg} is inherently unavailable, making predicted \acp{kg} the realistic standard for these comparisons.

\section{Discussion}

The core strength of \ac{biopie} lies in its procedure-centric, rather than concept-centric, design philosophy. In \ac{biopie}, experimental operations are treated as the fundamental units, and the structured dependencies among actions, objects, parameters, and procedural steps are explicitly captured. This design enables complex biomedical experiment understanding: our system achieves 83.33\% accuracy on the \ac{hid} questions and 74.80\% on the \ac{msr} questions, outperforming all baselines. These results demonstrate that \ac{biopie} can effectively process content with \ac{hid} and support \ac{msr}, highlighting its potential to serve as a foundation for experiment-level understanding. \ac{biopie} is also constructed at a practically reasonable scale. The diminishing improvements in Fig.~\ref{fig:iecompare}\textbf{(B)} indicate decreasing marginal returns from additional training data, with \ac{re} in particular exhibiting clear saturation behavior, while \ac{ner} shows only slow improvement with markedly diminishing returns. 

While closed-source \acp{llm} might have encountered the raw texts during pretraining, this potential exposure does not compromise our evaluation fidelity. Our benchmarks are explicitly designed to assess operational capabilities rather than textual memorization: first, the \ac{ie} task evaluates how effectively a model adapts to our newly defined, schema-constrained templates for generating fine-grained procedural \acp{kg}; second, the \ac{qa} system \sout{strictly} measures performance in formulating responses strictly grounded within structured evidence and local constraints.

Beyond standard \ac{qa} systems, \ac{biopie} can serve as a structured framework to analyze and decompose complex human instructions in the biomedical domain. By formalizing natural language protocols into structured representations, the dataset helps analyze intricate experimental logic and facilitates intent understanding.

As a structured human-robot interface, \ac{biopie} can assist in bridging the gap between human-readable protocols and robotic execution scripts. It facilitates basic workflow verification, such as local parameter consistency and constraint checking. Furthermore, the dataset offers a reusable semantic reference that can support modular protocol composition and conditional adaptation. Consequently, \ac{biopie} provides a valuable decision-making reference for \ac{llm}-based planners (\eg, ReAct~\citep{yao2022react}), contributing to more reliable integration between \ac{ai} assistants and robotic execution systems in automated laboratories (see Appx.~\ref{apxsec:qaapplication} for detailed discussion).

\section{Conclusion}

In this work, we investigate the problem of biomedical experiment \ac{qa} from the perspective of structured procedural understanding. We introduce \ac{biopie}, a new \ac{ie} dataset designed to capture fine-grained experimental entities, actions, and procedural relations while maintaining sufficient breadth to generalize across biomedical research. A comprehensive benchmark on \ac{biopie} indicates existing supervised models and \acp{llm} face challenges in protocol-centric \ac{ie}, particularly with \ac{ood} protocols. A \ac{qa} system is proposed to evaluate the \ac{qa} performance enhancement with \ac{biopie}. Both the \ac{qa} evaluation results and ablation studies highlight the crucial role of \ac{biopie} in complex reasoning, including \ac{hid} and \ac{msr}, of biomedical experiment protocols.

% \clearpage
\section*{Limitations}
Despite our efforts, constructing a gold-standard dataset for \ac{ie} over biomedical experimental protocols remains challenging. One limitation arises from our use of the \acp{llm} for protocol text normalization. While normalization improves consistency, it may introduce misalignments in step references, \eg, references to the product of an earlier step may be shifted to a later step after normalization. Such errors can affect fine-grained step-level grounding and temporal dependency annotation.

\faExclamationTriangle\ \textbf{Warning.} Reproducing the biomedical experiments described in \acs{biopie} \textbf{must only be carried out under the direct supervision of qualified domain experts}, as many procedures \textbf{involve significant safety hazards} and may \textbf{pose serious risks to personnel, equipment, and the environment} if performed improperly. The biomedical protocols provided are strictly for reference purposes only and are not intended to serve as standalone or executable experimental instructions. This is consistent with their presentation in the original publications.

\section*{Ethical Statement}
The original natural language descriptions are sourced from three websites, including Nature\hyperlink{fn:nature}{\textsuperscript{1}}, Cell\hyperlink{fn:cell}{\textsuperscript{2}}, and JoVE\hyperlink{fn:jove}{\textsuperscript{3}}. We further performed data cleaning and annotation on these descriptions. We carefully ensured that all experimental protocols incorporated into our corpus strictly comply with open-access policies and are distributed under Creative Commons licenses. This guarantees full adherence to copyright and intellectual property regulations, without any infringement or unauthorized use of protected materials.
\footnotetext[1]{\hypertarget{fn:nature}{}\url{https://www.nature.com/}}
\footnotetext[2]{\hypertarget{fn:cell}{}\url{https://star-protocols.cell.com/}}
\footnotetext[3]{\hypertarget{fn:jove}{}\url{https://www.jove.com/}}

\section*{Reproducibility}
Both the \ac{biopie} and \ac{qa} datasets are available at \url{https://sites.google.com/view/biopie}.

\section*{Acknowledgments}
This work was supported by the National Natural Science Foundation of China (Grant No. 52475001). The authors would like to thank Linkerbot Co., Ltd. for providing the dexterous robotic hand used in this study. The authors also thank Yu-Zhe Shi for assistance with early-stage data collection and Jiawen Liu for helpful discussions related to the figures.

\bibliography{references}

\appendix
\renewcommand\thefigure{A\arabic{figure}}
\setcounter{figure}{0}
\renewcommand\thetable{A\arabic{table}}
\setcounter{table}{0}
\renewcommand\theequation{A\arabic{equation}}
\setcounter{equation}{0}
\pagenumbering{arabic}% resets `page` counter to 1
\renewcommand*{\thepage}{A\arabic{page}}
\setcounter{footnote}{0}

\section{Details of \acs{biopie}}

\subsection{Development of the Annotation Scheme}\label{apxsec:ontologies}

\subsubsection{Scope of the Proposed Dataset}

Specifically, \ac{biopie} is bound to foundational laboratory explorations, rather than clinical cohort studies or medical interventions targeting human patients, with a particular emphasis on mainstream experimental setups and methodologies established between 2020 and 2025. The dataset is tailored for wet-lab workflows that involve physical interactions among reagents and materials, excluding bioinformatics data analyses (dry-lab). Accordingly, the experimental subjects represented in our dataset encompass cells, tissues, animal models, plants, and microorganisms—spanning both in vivo and in vitro paradigms for living systems—while excluding human subject groups. By concentrating our source corpus on operation-intensive branches such as genetic manipulation, molecular interactions, physiological signal detection, and regenerative medicine, \ac{biopie} models fine-grained laboratory procedures; however, while it aims to achieve broad applicability, generalizability across all peripheral biomedical disciplines cannot be guaranteed.

\subsubsection{Design Process}
Our scheme was developed through an iterative, corpus-driven process. We began with pilot annotations on a representative subset of experimental protocols, then refined entity and relation types through multiple rounds of adjudication between the annotators and domain experts. Each revision was motivated by observed annotation conflicts or coverage gaps rather than arbitrary design choices. Some entity or relation types were merged or split based on inter-annotator disagreement analysis, which is standard practice in corpus annotation methodology.

\subsubsection{Comparison with Existing Biomedical Ontologies}

The scheme was initially designed with reference to established formal ontologies for experimental procedures, namely EXACT2~\citep{soldatova2014exact2} and OBI~\citep{bandrowski2016ontology}. However, this alignment remains descriptive and ontology-inspired rather than operationally ontology-grounded. while EXACT2 and OBI provide machine-readable semantic representations of experimental procedures, our scheme is a text-span annotation scheme intended for information extraction from free-text protocols.

The correspondences are as follows:
\begin{itemize}
\item EXACT2's action participants (reagents, materials, instruments) correspond to our entity types \texttt{chemical}, \texttt{biomaterial}, \texttt{protein}, \texttt{nucleic acid}, \texttt{cell}, \etc.

\item EXACT2's quantitative parameter slots map directly to our \texttt{temperature}, \texttt{time}, \texttt{volume}, \texttt{concentration}, \texttt{speed}, \texttt{force}, \texttt{mass}, and \texttt{length} entity types.

\item OBI's material entity hierarchy (organism, anatomical part, cell line) informs our \texttt{organ}, \texttt{animal}, \texttt{plant}, and \texttt{cell} types.

\item At the relation level, EXACT2's \texttt{has\_object}, \texttt{has\_instrument}, \texttt{has\_reagent}, and \texttt{has\_parameter} slots correspond to our \texttt{is\_object\_of}, \texttt{use\_device}, \texttt{use\_reagent}, and \texttt{have\_parameter} relations, respectively. Action sequencing in EXACT2 maps to our \texttt{next\_step} relation.
\end{itemize}

We additionally introduce entity types (\texttt{software}, \texttt{data}, \texttt{position}, \texttt{blend}) and relation types (\texttt{during}, \texttt{for\_each}, \texttt{based\_on}, \texttt{in\_condition\_of}, \texttt{is\_goal\_of}, \etc) not prominently covered by EXACT2 or OBI, motivated by patterns frequently observed in our corpus. The scheme also underwent data-driven iterative refinement during annotation, some types were merged or split based on inter-annotator disagreement analysis, which is standard practice in corpus annotation.

\subsubsection{Recovering Investigation-level Information from Fine-grained Procedural Annotations}\label{apxsec:comprehensive}

\begin{table*}[htbp]
    \centering
    \small
    \renewcommand{\arraystretch}{1}
    \begin{tabularx}{\textwidth}{@{} >{\raggedright\arraybackslash}p{2.5cm} >{\raggedright\arraybackslash}X >{\raggedright\arraybackslash}p{3.5cm} >{\raggedright\arraybackslash}p{3.5cm} @{}}
        \toprule
        \textbf{Design / Focus} & \textbf{Original Protocol Evidence} & \textbf{Annotation Evidence (Examples)} & \textbf{Investigation-level Interpretation} \\
        \midrule
        
        \textbf{Dose--control--readout design} & 
        - ``Apply physical dose levels of 1 Gy and 0.5 Gy, resulting in expected carbon ion beam fluences of $8.43\times10^{6}$ cm$^{-2}$ and $4.16\times10^{6}$ cm$^{-2}$, respectively.'' \newline
        - ``Place potential stopping material (e.g., PMMA)...'' \newline
        - ``Shield control well positions with additional stopping material'' \newline
        - ``Ensure identical positions are used for correlation...'' \newline
        - ``Analyze RIF for single cells over time'' \newline
        - ``Quantify RIF number, size, and signal intensity.'' & 
        \texttt{["physical dose levels", "have\_property", "1 Gy"]} \newline
        \texttt{["Shield", "apply\_material", "additional stopping material"]} \newline
        \texttt{["RIF number", "is\_object\_of", "Quantify"]} & 
        Supports the reconstruction of two irradiation conditions, shielded control positions, dose-related variables, matched measurement settings, and single-cell RIF readouts associated with irradiation conditions. \\
        \addlinespace
        
        \textbf{Donor-to-recipient provenance} & 
        - ``Isolate spleens from CTL-donor mice'' \newline
        - ``Generate a single cell suspension by extruding the spleen...'' \newline
        - ``Inject 200 $\mu$l ($2.5\times10^{6}$ cells) intravenously into mice...'' \newline
        - ``Inject antigen with or without adjuvant intravenously into mice 16 hours after adoptive T cell transfer'' \newline
        - ``Isolate spleens from injected mice after 10 hours.'' \newline
        - ``Determine the number of labeled cells...'' & 
        \texttt{["Isolate", "from", "CTL-donor mice"]} \newline
        \texttt{["adoptive T cell transfer", "next\_step", "16 hours"]} \newline
        \texttt{["number of labeled cells", "is\_object\_of", "Determine"]} & 
        Maps the provenance chain (CTL-donor $\rightarrow$ spleen $\rightarrow$ adoptive transfer $\rightarrow$ recipient spleen), compared treatment conditions, sampling times, and quantitative readouts. \\
        \addlinespace
        
        \textbf{Genetic perturbations \& mechanisms} & 
        - ``Electrophorese lab-purified recombinant proteins MITOL (WT and CD) in SDS-PAGE'' \newline
        - ``Determine the exact concentrations of MITOL WT and MITOL CD...'' \newline
        - ``Add 1 $\mu$g Flag-Pol$\gamma$A, 1 $\mu$g Myc-MITOL WT, 1 $\mu$g MITOL CD, and 100 ng His-Ubiquitin plasmids...'' \newline
        - ``Analyze ubiquitylation of substrate protein by Western blot...'' \newline
        - ``Determine that equal amounts of Pol$\gamma$A... were used'' & 
        \texttt{["recombinant proteins MITOL", "contain", "WT"]} \newline
        \texttt{["Analyze", "use\_method", "Western blot"]} \newline
        \texttt{["exact concentrations...", "is\_object\_of", "Determine"]} & 
        Preserves manipulated MITOL catalytic states, alternative Pol$\gamma$A substrate variants, calibrated protein inputs, an equal-input check, and the ubiquitylation readout. \\
        \addlinespace
        
        \textbf{Animal treatment \& tissue lineage} & 
        - ``Adjust the airflow to deliver a mixture of 2.3\% sevoflurane...'' \newline
        - ``Continuously deliver sevoflurane, medical air, and oxygen for 2 hours.'' \newline
        - ``Quickly extract the brain and keep it in ice-cold carbonated aCSF'' \newline
        - ``Cut coronal slices...'' \newline
        - ``Carefully remove hippocampal slices from the coronal slices...'' \newline
        - ``Load 10 $\mu$L of each protein fraction in the same gel for comparison by Western blotting'' & 
        \texttt{["mixture", "contain", "sevoflurane"]} \newline
        \texttt{["hippocampal slices", "is\_object\_of", "remove"]} \newline
        \texttt{["comparison", "use\_method", "Western blotting"]} & 
        Connects animal-level anesthetic exposure to the lineage (treated mouse $\rightarrow$ brain $\rightarrow$ coronal slices $\rightarrow$ hippocampal slices) and matched protein fraction comparisons. \\
        \addlinespace
        
        \textbf{Comparative methods \& statistics} & 
        - ``Blend cells into the bioink using the passive mixing unit technique 1, 2, or 3 times...'' \newline
        - ``Blend cells into a separate bioink using manual mechanical mixing...'' \newline
        - ``Isolate and cryopreserve primary human nasal chondrocytes (hNC) from patients'' \newline
        - ``Calculate cell viability based on the ratio of live cells...'' \newline
        - ``Analyze the data via one-way ANOVA followed by Tukey's multiple comparisons test'' & 
        \texttt{["Blend", "repetitions", "1"]} \newline
        \texttt{["primary human nasal chondrocytes", "from", "patients"]} \newline
        \texttt{["Analyze", "use\_method", "one-way ANOVA"]} & 
        Exposes comparison groups by mixing method, patient-derived cell provenance, quantitative definition for cell viability, inferential analysis, and longitudinal endpoints. \\
        
        \bottomrule
    \end{tabularx}
    \caption{\textbf{Original protocol sentences and selected relation annotations supporting the corresponding investigation-level interpretations.}}
    \label{tab:protocol_summary_original}
\end{table*}

\ac{biopie} is designed as a procedure-centric information-extraction scheme rather than a complete investigation ontology. Nevertheless, this does not mean that investigation-level information is absent from the annotations. When such information is explicitly stated in a protocol, it can often be compositionally recovered from the serialized local steps, because \ac{biopie} jointly preserves actions, biological and non-biological participants, quantitative parameters, typed semantic relations, and procedural order. In particular, alternative materials, dosages, variants, and processing conditions provide evidence for experimental designs and treatment/control arms; \texttt{from}, \texttt{to}, and step-sequencing relations preserve biosample trajectories; measurement actions and data-bearing entities identify outcome variables and analysis procedures; and relations such as \texttt{based\_on}, \texttt{is\_goal\_of}, \texttt{during}, and \texttt{next\_step} connect interventions, observations, and downstream readouts.

Tab.~\ref{tab:protocol_summary_original} quotes the original protocol sentences in the dataset and show selected relation annotations that support the corresponding investigation-level interpretations.

These examples clarify the intended scope of \ac{biopie}. The scheme directly represents protocol actions, participants, instruments, quantities, conditions, local semantic relations, and procedural dependencies. From these annotations, treatment and control conditions, biosample trajectories, manipulated variables, measurement variables, time points, comparison structures, and intervention--readout evidence chains can be reconstructed when they are explicitly described in the source text. However, \ac{biopie} does not normalize these elements into dedicated global objects such as \texttt{ExperimentalDesign}, \texttt{ControlArm}, \texttt{Outcome}, or \texttt{Mechanism}, nor does it aim to encode complete hypotheses, cohort-level metadata, randomization or blinding procedures, raw measurement matrices, statistical results, causal conclusions, or formal biological-mechanism graphs unless these are directly stated as procedural content.

Accordingly, \ac{biopie} should be understood as a procedure-grounded intermediate IE layer that is complementary to OBI and EXACT2. Its dense text-span annotations provide grounded entities and relations that can support downstream aggregation into investigation-level representations, whereas formal ontologies provide the global conceptual structures that \ac{biopie} intentionally does not duplicate. We therefore use ``biomedical experiment understanding'' to refer specifically to understanding how biomedical experiments are operationally executed, conditioned, sampled, and measured, rather than to a complete formal representation of every dimension of a biomedical investigation.

\subsubsection{Complementarity with the Experimental Factor Ontology}

The \ac{efo} provides standardized concepts for experimental variables, such as biological materials, chemical treatments, assays, and measured phenotypes~\citep{malone2010modeling}. \ac{biopie} is complementary to \ac{efo}: \ac{efo} supports concept normalization, while \ac{biopie} captures how these variables are used in protocol steps through actions, parameters, semantic relations, and procedural order.

\begin{itemize}
\item \textbf{Paired treatment and vehicle control.}

\textit{Original protocol text:} ``Topically apply PMA (100 $\mu$M, 20 $\mu$L) to the left ear of each mouse.'' ``Topically apply vehicle (ethanol, 20 $\mu$L) to the right ear of each mouse.'' ``Measure the photon fluxes in each ROI.''

\textit{Complementarity:} \ac{efo}-compatible concepts can normalize PMA, ethanol, ear anatomy, and photon flux. \ac{biopie} additionally links each material to its application action, dose, volume, anatomical site, and downstream measurement. These relations support reconstruction of the within-animal treatment--control design.

\item \textbf{Cell type and electrophysiological measurement conditions.}

\textit{Original protocol text:} ``Load 1 mL of Fluo-4-AM working solution (2 $\mu$M) to human DA neurons.'' ``Hold the cell at $-60$ mV in voltage-clamp configuration.'' ``Record Ca$^{2+}$ currents by stepping from $-60$ mV to $+50$ mV for 250 ms in 10 mV increments.''

\textit{Complementarity:} \ac{efo} and related ontologies can normalize the cell type, chemical probe, assay, and measured variable. \ac{biopie} preserves the loading concentration and volume, target cells, holding potential, voltage range, pulse duration, increment, and Ca$^{2+}$-current readout.

\item \textbf{Stage-specific differentiation factors.}

\textit{Original protocol text:} ``When human iPS cell colonies reach 80\% confluence, DE induction is performed.'' ``Mix RPMI medium/Glutamax with Activin A (100 ng/mL) and Wnt 3a (25 ng/mL).'' ``Mix DMEM/F12 basal medium with Noggin (200 ng/mL) and SB-431542 (10 $\mu$M).'' ``Mix DMEM/F12 basal medium with FGF2 (250 ng/mL), FGF7 (100 ng/mL), and FGF10 (100 ng/mL).''

\textit{Complementarity:} \ac{efo}-compatible terms can normalize the cell type, differentiation process, growth factors, and inhibitors. \ac{biopie} additionally preserves the induction condition, exact concentrations, factor combinations, and their positions in successive differentiation stages.
\end{itemize}

These examples show that \ac{efo} and \ac{biopie} operate at complementary levels. \ac{efo} provides standardized identifiers and concept hierarchies, whereas \ac{biopie} provides text-grounded procedural context, including actions, targets, parameters, and step order.

\subsection{Data Source}\label{apxsec:datasource}

\begin{table}[t]
\centering
\resizebox{0.75\linewidth}{!}{%
\begin{tabular}{lcccc}
\toprule
{Source} & {Train} & {Dev} & {\acs{id} Test} & {\acs{ood} Test}\\
\midrule
Cell   & 115 & 16 & 32 & 0  \\
Nature & 82  & 9  & 20 & 17 \\
JoVE   & 43  & 11 & 16 & 28  \\
\bottomrule
\end{tabular}
}
\caption{\textbf{Distribution and source journals of the \acs{biopie} protocols.}}
\label{tab:biopie_origin}
\end{table}

\begin{table}[t]
\centering
\resizebox{\linewidth}{!}{%
\begin{tabular}{lcccc}
\toprule
& {Train} & {Val} & {\ac{id}} & {\ac{ood}} \\
\midrule
Genetic Manipulation & 60 & 14 & 28 & 0 \\
Molecular Interactions & 73 & 4 & 13 & 0 \\
Physiological Signal Detection & 55 & 8 & 15 & 0 \\
Regenerative Medicine & 52 & 10 & 12 & 0 \\
\midrule
Plant-based Biopharmaceuticals & 0 & 0 & 0 & 28 \\
Functional Imaging & 0 & 0 & 0 & 10 \\
Epidemiological Analysis & 0 & 0 & 0 & 7 \\
\bottomrule
\end{tabular}
}
\caption{\textbf{Domain distribution and dataset splits in \acs{biopie} protocols.} The dataset includes four primary domains for training and \ac{id} testing, and three specialized domains reserved exclusively for \ac{ood} evaluation to test model robustness across diverse biomedical fields.}
\label{tab:domain_stats}
\end{table}

To construct \ac{biopie}, we curated a corpus of biological protocols from three primary high-quality sources: Cell, Nature, and JoVE, as shown in Tab. \ref{tab:biopie_origin}. We targeted protocols published between 2020 and 2025. Our sampling strategy was designed to cover a wide range of biological sub-domains, as shown in Tab. \ref{tab:domain_stats}.

\subsection{The Impact of Normalization on Protocols}\label{apxsec:normalization}

\subsubsection{Qualitative Analyses}

The raw protocol text sourced from Nature Protocols is often written in a mixed style—containing compound sentences, parenthetical reagent details, conditional instructions, passive constructions, and non-actionable commentary. The normalization step, powered by Qwen-max, decomposes and rewrites this text into discrete, imperative, subject-free sentences, each representing a single atomic action. This standardized format is essential for downstream annotation tasks such as action-type labeling and argument extraction.

\textit{Example 1: Confocal Microscopy}
\begin{lstlisting}[]
Raw text:
"Acquire Z-sections using a custom-made spinning disk confocal microscope as described above."
Normalized output:
"Acquire Z-section images using a custom-made spinning disk confocal microscope."
\end{lstlisting}
Here the normalization removes the vague cross-reference ("as described above") while preserving the core imperative action, making the step self-contained and annotation-ready.

\textit{Example 2: Haplotype-Fusion PCR }
\begin{lstlisting}[]
Raw text:
"Transfer completed PCR reactions to a 1.5 ml tube and add 400 µl hexane. Vortex thoroughly and centrifuge at full speed for 3 minutes. Remove and discard hexane layer (upper layer). Repeat with 400 µl fresh hexane."
Normalized output (4 discrete steps):
"Transfer completed PCR reactions to a 1.5 ml tube and add 400 µl hexane."
"Vortex thoroughly and centrifuge at full speed for 3 minutes."
"Remove and discard the upper hexane layer."
"Repeat with 400 µl fresh hexane."
\end{lstlisting}
This example illustrates the most common normalization operation: splitting compound sentences joined by conjunctions into the separate atomic steps. The parenthetical clarification "(upper layer)" is also integrated naturally into the imperative sentence rather than left as a parenthetical.

\textit{Example 3: Gap1 Integrative Vector}
\begin{lstlisting}[]
Raw text:
"Amplify a pUC fragment containing the bacterial origin of replication and the kanamycin resistance marker by PCR from PCR-BluntII-TOPO (Invitrogen, Carlsbad, CA) and ligate with a PvuII fragment of pDRf1 containing the f1 origin of replication, the PMA1 promoter and ADH3 terminator producing pDL001."
Normalized output (2 discrete steps):
"Amplify a pUC fragment containing the bacterial origin of replication and the kanamycin resistance marker by PCR from PCR-BluntII-TOPO."
"Ligate the amplified pUC fragment with a PvuII fragment of pDRf1 to produce pDL001."
\end{lstlisting}
Here, the normalization does two things simultaneously: it separates two sequential actions (amplify, then ligate) that were conjoined in one sentence, and it removes the vendor information ("Invitrogen, Carlsbad, CA") which is not part of the procedural action itself. The second normalized step also adds a clarifying reference ("the amplified pUC fragment") to preserve logical coherence between steps.

\textit{Example 4: Simplified Behavioural Tests}
\begin{lstlisting}[]
Raw text:
"Begin the first trial by introducing a stimulus female into the home cage of a male mouse for a 1-min interaction. Record investigation time of the female by the male mouse. At the end of the 1-min trial, remove the stimulus animal and return to an individual holding cage. Score behaviour during experiments and record behaviour on a videotape. Scoring should be done by trained raters (inter-rater reliability >90%)."
Normalized output:
"Introduce a stimulus female into the home cage of a male mouse for a 1-minute interaction."
"Remove the stimulus animal after the 1-minute trial and return the male to an individual holding cage."
"Score behavior during the experiment and record on videotape."
\end{lstlisting}
This example demonstrates that normalization also filters out non-procedural meta-commentary ("Scoring should be done by trained raters with inter-rater reliability >90\%") that describes quality criteria rather than an executable action, keeping only genuinely actionable steps.

\subsubsection{Quantitative Analyses}

\begin{table}[t]
\centering
\resizebox{\linewidth}{!}{%
\begin{tabular}{lcc}
\toprule
& Action Verbs & Descriptive Verbs \\
\midrule
Original & 135 & 52 \\
Normalized & 125 & 11\\
Retention Rate & 92.59\% & 21.15\% \\
\bottomrule
\end{tabular}
}
\caption{\textbf{Quantitative comparison of verb distribution between the original and normalized protocols.}}
\label{tab:contentcom}
\end{table}

\begin{table}[t]
\centering
\resizebox{0.75\linewidth}{!}{%
\begin{tabular}{lccc}
\toprule
& Test & \ac{hid} & \ac{msr}\\
\midrule
Original   & 60.90 & 68.26 & 53.66 \\
Normalized & 61.52 & 66.09 & 56.10 \\
\bottomrule
\end{tabular}
}
\caption{\textbf{\ac{qa} performance evaluation for text normalization.}}
\label{tab:semanticcom}
\end{table}

To verify that the transformation of raw protocols into normalized, stepwise imperative sentences preserves the integrity of experimental information, we evaluate the normalization process from two perspectives: content fidelity and semantic consistency.

\textbf{Quantitative Impact on Information Content.}
We randomly sampled 102 pairs (original vs. normalized) and manually annotated the occurrence of two functional verb types: (1) action verbs, which denote executable experimental operations (\eg, centrifuge, incubate, add); (2) descriptive verbs, which primarily explain experimental principles, background, or non-procedural context (\eg, is, are, recommend).

The results (Table \ref{tab:contentcom}) indicate that the normalized text retains 92.59\% of the action verbs, ensuring that the core procedural logic remains intact. Manual inspection reveals that this marginal reduction primarily stems from the consolidation of redundant operations and the exclusion of verbs (\eg, ensure, check) that do not alter the core experimental execution. Conversely, there is a significant reduction in descriptive verbs, suggesting that the normalization process primarily impacts descriptive content. By filtering out redundancy, the process effectively distills the text while maintaining the essential operational density required for precise reasoning. Our analysis also shows a 98.08\% retention rate for numerical parameters and reagents, suggesting that safety-critical facts are preserved.

\textbf{Impact on Semantic Representation.} 
To further assess whether normalization preserves the core operational logic, we conducted a comparative analysis using our \ac{qa} dataset. We employed a text embedding model OpenAI's \textit{text-embedding-3-large} as a retriever to construct a vector database from both the raw and the normalized corpora.

Experimental results demonstrate that the \ac{qa} system leveraging normalized text maintains and exceeds the performance of the raw text baseline in several cases (Tab. \ref{tab:semanticcom}). This leads to two conclusions: 

The imperative structure used in normalization maintains a high degree of procedural fidelity to the original descriptive prose, allowing embedding models to capture core experimental facts accurately.

By eliminating linguistic noise and standardizing sentence structures, normalization enhances the signal-to-noise ratio. This assists the retriever in identifying precise experimental parameters (\eg, specific temperatures or durations), thereby providing cleaner evidence for \ac{hid} and \ac{msr} reasoning. The slight decline on the HID set is not statistically significant ($p \geq$ 0.0625 > 0.05), indicating minor variance rather than a meaningful loss of semantic information.

\begin{table*}[t]
\centering
\resizebox{0.80\textwidth}{!}{%
\begin{tabular}{lcccccccc}
\toprule
\multirow{2}{*}{} & \multicolumn{4}{c}{In-domain} & \multicolumn{4}{c}{Out-of-domain} \\
& \ac{ner} & Rel & Rel+ & \ac{re} & \ac{ner} & Rel & Rel+ & \ac{re} \\
\midrule
\multicolumn{9}{c}{\textit{Supervised Baselines}} \\
\midrule
PL-Marker$^1$~\citep{ye2022packed} & \textbf{87.99} & \textbf{83.99} & \textbf{75.82} & \textbf{88.72} & 77.24 & \textbf{75.83} & \textbf{56.40} & \textbf{80.81} \\
PL-Marker$^2$~\citep{ye2022packed} & 86.87 & 81.97 & 73.84 & 80.14 & \textbf{88.24} & 70.85 & 49.78 & 71.79 \\
HGERE$^1$~\citep{yan2023joint}     & 87.01 & 82.92 & 74.68 &  -  & 74.85 & 72.59 & 53.97 &  -  \\
HGERE$^2$~\citep{yan2023joint}     & 86.93 & 81.79 & 73.86 & - & 73.76 & 72.36 & 53.62 & - \\

\midrule
\multicolumn{9}{c}{\textit{Zero-shot \ac{llm}}} \\
\midrule
GPT-5 (Pipeline)         & 47.61 & 47.83 & 39.00 & 64.81 & 41.96 & 47.98 & 35.06 & 65.14 \\
GPT-5 (Joint)            & 45.89 & 61.74 & 48.70 &  -   & 50.28 & 61.09 & 42.53 &  -   \\
Claude-4.5-Opus (Pipeline)    & 71.55 & 41.87 & 32.93 & 49.73 & 66.28 & 35.31 & 25.58 & 45.45 \\
Claude-4.5-Opus (Joint)       & 71.31 & 35.11 & 27.23 &  -   & 67.76 & 32.55 & 25.77 &  -   \\
Llama-4 (Pipeline)       & 31.86 & 7.14  & 5.46  & 16.67 & 32.91 & 8.82  & 6.86  & 13.33 \\
Llama-4 (Joint)          & 58.62 & 23.08 & 0.00  &  -   & 54.55 & 28.57 & 14.29 &  -   \\
Qwen-max (Pipeline)      & 70.28 & 22.82 & 18.93 & 29.69 & 64.77 & 20.90 & 15.05 & 27.61 \\
Qwen-max (Joint)         & 69.34 & 26.39 & 20.83 &  -   & 66.64 & 22.29 & 15.64 &  -   \\
\midrule
\multicolumn{9}{c}{\textit{Few-shot \ac{llm}}} \\
\midrule
GPT-5 (Pipeline)         & 51.12 & 62.89 & 56.20 & 75.80 & 42.33 & 51.15 & 39.23 & 62.92 \\
GPT-5 (Joint)            & 54.20 & 65.50 & 57.95 &  -   & 49.55 & 63.06 & 51.11 &  -   \\
Claude-4.5-Opus (Pipeline)    & 85.62 & 68.67 & 63.29 & 77.25 & 75.82 & 53.88 & 41.86 & 65.35 \\
Claude-4.5-Opus (Joint)       & 83.89 & 66.01 & 60.38 &  -   & 72.28 & 53.11 & 41.75 &  -   \\
Llama-4 (Pipeline)       & 35.81 & 18.82 & 16.83 & 21.12 & 30.10 & 10.69 & 8.88  & 14.81 \\
Llama-4 (Joint)          & 53.57 & 32.09 & 30.70 &  -   & 57.09 & 26.40 & 20.00 &  -   \\
Qwen-max (Pipeline)      & 85.14 & 65.67 & 61.11 & 73.52 & 74.94 & 48.63 & 38.34 & 57.86 \\
Qwen-max (Joint)         & 84.68 & 61.80 & 57.73 &  -   & 72.85 & 43.72 & 34.36 &  -   \\
\midrule
\multicolumn{9}{c}{\textit{\ac{lora} \ac{llm}}} \\
\midrule
Llama-3-8B (Pipeline)    & 86.54 & 75.45 & 68.29 & 82.52 & 77.38 & 64.44 & 50.60 & 73.70 \\
Llama-3-8B (Joint)       & 84.73 & 75.52 & 67.16 &  -   & 76.80 & 64.58 & 48.13 &  -   \\
Qwen-3-8B (Pipeline)     & 84.70 & 70.65 & 63.31 & 78.14 & 77.23 & 64.07 & 48.67 & 69.15 \\
Qwen-3-8B (Joint)        & 83.38 & 72.20 & 64.30 &  -   & 76.42 & 62.97 & 46.76 &  -   \\
\bottomrule
\end{tabular}
}
\caption{\textbf{Test precision scores of different baselines on our proposed dataset.}  $^1$ Denotes the use of \textit{scibert}~\citep{beltagy2019scibert} as the encoder; $^2$ denotes \textit{biobert}~\citep{lee2020biobert}. } 
\label{tab:ieprebenchmark}
\end{table*}

\begin{table*}[t]
\centering
\resizebox{0.80\textwidth}{!}{%
\begin{tabular}{lcccccccc}
\toprule
\multirow{2}{*}{} & \multicolumn{4}{c}{In-domain} & \multicolumn{4}{c}{Out-of-domain} \\
& \ac{ner} & Rel & Rel+ & \ac{re} & \ac{ner} & Rel & Rel+ & \ac{re} \\
\midrule
\multicolumn{9}{c}{\textit{Supervised Baselines}} \\
\midrule
PL-Marker$^1$~\citep{ye2022packed} & 86.82 & 81.16 & \textbf{73.27} & \textbf{87.08} & 70.78 & 65.48 & 48.71 & \textbf{76.98} \\
PL-Marker$^2$~\citep{ye2022packed} & 85.80 & 78.99 & 71.16 & 21.32 & 36.80 & 14.09 & 9.90 & 14.99 \\
HGERE$^1$~\citep{yan2023joint}     & \textbf{88.25} & \textbf{81.29} & 73.20 & - & \textbf{74.31} & \textbf{68.51} & \textbf{50.94} & - \\
HGERE$^2$~\citep{yan2023joint}     & 86.97 & 80.08 & 72.31 & - & 71.79 & 66.82 & 49.51 & - \\
\midrule
\multicolumn{9}{c}{\textit{Zero-shot \ac{llm}}} \\
\midrule
GPT-5 (Pipeline)         & 71.44 & 53.41 & 43.54 & 75.75 & 68.62 & 55.47 & 40.53 & 71.65 \\
GPT-5 (Joint)            & 15.26 & 13.56 & 10.69 &   -   & 14.03 & 12.84 & 8.94  &   -   \\
Claude-4.5-Opus (Pipeline)    & 67.27 & 39.34 & 30.94 & 46.40 & 61.53 & 33.49 & 24.26 & 41.39 \\
Claude-4.5-Opus (Joint)       & 67.53 & 32.34 & 25.08 &   -   & 63.84 & 30.16 & 23.88 &   -   \\
Llama-4 (Pipeline)       & 57.78 & 1.08  & 0.83  & 0.89  & 59.37 & 0.86  & 0.67  & 0.38  \\
Llama-4 (Joint)          & 0.88  & 0.38  & 0.00  &   -   & 0.46  & 0.19  & 0.10  &   -   \\
Qwen-max (Pipeline)      & 64.23 & 18.65 & 15.47 & 25.40 & 56.28 & 15.98 & 11.51 & 22.93 \\
Qwen-max (Joint)         & 62.47 & 22.34 & 17.63 &   -   & 56.82 & 18.17 & 12.75 &   -   \\
\midrule
\multicolumn{9}{c}{\textit{Few-shot \ac{llm}}} \\
\midrule
GPT-5 (Pipeline)         & 81.19 & 71.29 & 63.72 & 83.13 & 70.16 & 57.18 & 43.86 & 75.07 \\
GPT-5 (Joint)            & 18.61 & 15.47 & 13.69 &   -   & 25.75 & 21.60 & 17.51 &   -   \\
Claude-4.5-Opus (Pipeline)    & 84.74 & 69.06 & 63.65 & 77.15 & 71.70 & 52.90 & 41.10 & 64.41 \\
Claude-4.5-Opus (Joint)       & 82.94 & 65.75 & 60.15 &   -   & 71.78 & 52.05 & 40.91 &   -   \\
Llama-4 (Pipeline)       & 77.84 & 18.65 & 16.68 & 20.94 & 65.38 & 12.94 & 10.75 & 17.98 \\
Llama-4 (Joint)          & 7.73  & 4.39  & 4.20  &   -   & 10.87 & 6.28  & 4.76  &   -   \\
Qwen-max (Pipeline)      & 81.49 & 63.08 & 58.69 & 72.82 & 68.93 & 44.05 & 34.73 & 55.66 \\
Qwen-max (Joint)         & 68.40 & 50.35 & 47.04 &   -   & 54.20 & 32.45 & 25.50 &   -   \\
\midrule
\multicolumn{9}{c}{\textit{\ac{lora} \ac{llm}}} \\
\midrule
Llama-3-8B (Pipeline)    & 86.13 & 75.11 & 67.98 & 80.84 & 74.09 & 61.56 & 48.33 & 72.79 \\
Llama-3-8B (Joint)       & 84.69 & 74.22 & 66.00 &   -   & 73.01 & 60.89 & 45.39 &   -   \\
Qwen-3-8B (Pipeline)     & 84.18 & 69.26 & 62.06 & 76.00 & 72.71 & 59.37 & 45.10 & 68.03 \\
Qwen-3-8B (Joint)        & 82.47 & 69.26 & 61.68 &   -   & 71.47 & 57.28 & 42.53 &   -   \\
\bottomrule
\end{tabular}
}
\caption{\textbf{Test recall scores of different baselines on our proposed dataset.}  $^1$ Denotes the use of \textit{scibert}~\citep{beltagy2019scibert} as the encoder; $^2$ denotes \textit{biobert}~\citep{lee2020biobert}. } 
\label{tab:ierecbenchmark}
\end{table*}

\subsection{Annotation Guideline, Data Scheme Definition, and \acs{ie} Prompt}\label{apxsec:guideline}

This section provides the annotation guidelines for the proposed dataset, covering the data scheme definition and the procedures used for consistent annotation.

The prompt for \ac{llm}-based joint extraction is the guideline shown below, while the prompt for pipeline extraction is obtained by splitting the following prompt.

\begin{lstlisting}[]
You are given a piece of text describing biomedical experiments or laboratory workflows.
Your task is to identify all factual entities and all relationships between these entities.

The possible entity types are listed below.
- verb: Actions performed in a procedure. (e.g., Fix, Osmicate, Dehydrate)
- part: Specific sections of an object. (e.g., upper surface of the specimen, plunger, plunger of the bioink syringe)
- container: Objects used to hold substances. (e.g., original culture plate, cartridge, well plate)
- force: Physical force or weight applied. (e.g., 500 g, 17,000 × g, 226 × g)
- device: Tools used in experiments. (e.g., fume hood, aluminum stub, underlying aluminum stub)
- method: Techniques for conducting experiments. (e.g., simultaneously, direct, trypan blue exclusion method)
- chemical: Substances used in a process excluding proteins and polymers (e.g., TAG, Karnovsky, aqueous osmium tetroxide)
- concentration: Ratio of a substance in a solution. (e.g., 1% (wt/vol), 50%, 70%)
- consumable: Materials used up in experiments. (e.g., sticky sellotape tabs, copper tape, silver paint)
- state: Condition of a material or system. (e.g., continuous contact, recorded, sterile)
- volume: Measurement of liquid quantity. (e.g., volumes, 12 mL, 1 mL)
- temperature: Heat level in a process. (e.g., room temperature, 4°C)
- time: Duration of an action or waiting. (e.g., 2 hours, overnight)
- process: Series of actions in a procedure. (e.g., air dry, cross-linking, additional blends)
- times: Number of repetitions. (e.g., three times, two, 1)
- cell: Basic biological unit in living organisms. (e.g., cell monolayers, samples, sample)
- nucleic acid: DNA or RNA sequences used in biological experiments. (e.g., genomic DNA, T7-RT primer, first-strand cDNA)
- biomaterial: Biological substances in use. (e.g., bioink)
- software: Programs for analysis or instrument control. (e.g., SmartSEM software, Nikon Imaging Software)
- number: Countable values in a process. (e.g., two, total number of cells)
- energy: Measure of work or electrical energy. (e.g., 3-5 KV, 400 mJ)
- speed: Rate of motion or process. (e.g., controlled rate, 20 rpm)
- mass: Quantity of matter. (e.g., final cell density, 2 µg)
- environment: Conditions affecting an experiment. (e.g., dust-free environment, standard conditions)
- length: Measurement of distance. (e.g., approximately 1 nm, working distance of 4 mm)
- data: Recorded experimental information. (e.g., TIFF images, digital image files)
- organ: Biological structures in research. (e.g., spleen, spleens)
- animal: Living organisms in studies. (e.g., mice, CTL-donor mice)
- protein: Functional biomolecules. (e.g., trypsin/EDTA solution, BSA/PBS solution)
- polymer: Large molecular compounds. (e.g., nanocellulose/alginate, agarose gel)
- position: Spatial location of an object or material. (e.g., in the printed construct, on the dispensing unit)
- size: Dimensional magnitude of an object. (e.g., approximate size of the plate, 220 × 220)
- plant: Botanical specimens or components used in experiments. (e.g., red beet, spinach)
- blend: Mixed substances. (e.g., bioink-cell mixture, blend, cell/bioink)

The possible relation types are listed below.
- is_object_of: Describes that an object is the target of an action. (e.g., cell monolayers is_object_of Fix)
- contain: Indicates that something contains another thing. (e.g., Zeiss Sigma microscope contain in-lens SE1 electron detector)
- use_method: Specifies the method used for an action. (e.g., Dehydrate use_method incubating)
- use_device: Specifies the device or tool used for an action. (e.g., air dry use_device fume hood)
- use_reagent: Specifies the reagent or chemical used in an action. (e.g., Fix use_reagent TAG)
- have_property: Describes a property of an object. (e.g., aqueous osmium tetroxide have_property 1% (wt/vol))
- apply_material: Specifies a material applied during an action. (e.g., stick apply_material sticky sellotape tabs)
- is_goal_of: Describes that a goal is the purpose of an action. (e.g., make is_goal_of Use)
- for_each: Specifies that an action applies to each specific object. (e.g., Place for_each sample)
- next_step: Indicates the next step after an action or process.  (e.g., 50% next_step 70%)
- to: Container or position to which an object or solution is transferred. (e.g., stick to aluminum stub)
- or: Represents alternative options. (e.g., TAG or Karnovsky)
- have_parameter: Specifies an action's or process's parameter. (e.g., Fix have_parameter room temperature)
- repetitions: Indicates the number of times an action is repeated. (e.g., Blend repetitions 1)
- use_software: Specifies software used. (e.g., Acquire use_software SmartSEM software)
- from: Indicates the source of something. (e.g., specimens from original culture plate)
- in_condition_of: Specifies the condition under which an action occurs. (e.g., Acquire in_condition_of 3-5 KV)
- not: Denotes negation or exclusion. (e.g., Mix not cartridge)
- during: Indicates that an event happens within the time frame of another. (e.g., Balance during choosing)
- equal: Expresses equivalence between two values or objects. (e.g., one equal syringes)
- based_on: Indicates dependence or derivation from something. (e.g., Calculate based_on total number of constructs desired)

The following rules define the annotation standards for Named-Entity Recognition (NER) and Relation Extraction (RE) in this dataset. Annotators should strictly adhere to these guidelines to ensure consistency and reproducibility.

General Principles
1. All annotations should preserve the original surface form as it appears in the text, without normalization or correction.
2. When uncertainty exists, prioritize precision over recall and omit questionable annotations rather than guessing.

Named-Entity Recognition (NER)
3. For NER, annotate all entity mentions and output only entity category pairs, one per line, in the following format:
```
entity: category
```
4. The entity span must be minimal and precise. Do not include determiners or function words such as "the", "a", or "this" within the entity span.
5. When both a full name and its abbreviation appear in the text, annotate each occurrence separately as independent entities.
6. Annotate every occurrence of an entity in the text, even if the same entity appears multiple times.
7. If an entity mention is ambiguous, assign the category that is most directly supported by the local context.
8. Overlapping or nested entity spans are permitted when they correspond to valid and distinct entity mentions.

Relation Extraction (RE)
9. For RE, annotate only explicitly stated or clearly implied relationships and output only relation triplets, one per line, in the following format:
```
head: head_entity  tail: tail_entity  relation: relationship
```
10. Both the head and tail entities must be annotated entity mentions present in the text.
11. Do not infer, assume, or hallucinate relations that are not directly supported by the text.
12. If multiple relations are expressed between the same entity pair, annotate each relation separately.
13. If the same relation involves an entity that appears in multiple positions in the text (e.g., via pronouns, or alternative mentions), annotate the relation only for the most salient or primary occurrence of that entity.
\end{lstlisting}

\subsection{Detailed \acs{ie} Benchmark Results}\label{apxsec:detailie}

Precision in the \ac{ie} benchmark is listed in Tab. \ref{tab:ieprebenchmark}. Recall in the \ac{ie} benchmark is listed in Tab. \ref{tab:ierecbenchmark}.

\section{\acs{qa} System Evaluation}

\subsection{\acs{qa} Dataset}\label{apxsec:qadatasetdetail}

We extracted 3275 sub-protocols from the complete collection of textual protocols (excluding those already structured as part of the \ac{ie} dataset) and automatically constructed the corresponding \ac{qa} pairs; the construction process is detailed in the Appx. \ref{apxsec:qadataset}. The dataset is divided into training, validation, and test sets with sizes of 1983, 159, and 1133, respectively. The training set is used for model training, the validation set is utilized for model hyper-parameter selection (the number of few-shot examples), and the test set is employed to evaluate the performance of the model. 

To further analyze model performance under challenging conditions, we construct two subsets from the test set. The first subset consists of 230 \ac{hid} questions, which are the questions in the test set with the top 230 highest relation counts. For these questions, the corresponding sentences from which they were generated contain an average of 10.40 relations, substantially higher than the overall average of 4.62 reported in Tab.~\ref{tab:dataset}. The second subset comprises 123 \acf{msr} questions (the average number of reasoning hops is 2.22), which includes all questions in the test set requiring more than one reasoning step. 

\subsection{\acs{qa} Dataset Construction}\label{apxsec:qadataset}

The \ac{biopie} \ac{qa} dataset was constructed automatically to validate the utility of the \ac{ie} dataset in supporting complex reasoning tasks while maintaining evaluation fairness. The process followed three rigorous steps:

\textbf{Step 1 \ac{kg}-based Question Generation}: We utilized the annotated \acp{kg} as logical skeletons. An \ac{llm} was used to generate natural language questions by identifying key entities and their target answers within the triplet structures.

\textbf{Step 2 Reasoning Path Refinement}: To ensure logical integrity, we screened the reasoning paths. If entities from the reasoning path appeared directly in the initial question, we adjusted the path or question to guarantee the correctness of the answer and the reasoning process.

\textbf{Step 3 Human Verification}: To ensure the quality, practical relevance, and independence of the validation, we implemented a comprehensive phase of human verification. Every single auto-generated question underwent manual screening by human annotators with expertise in biomedical fields. We filtered out any ambiguous, unreasonable, or erroneous content to ensure that the benchmark strictly aligns with the genuine needs, logic, and expectations of real-world biomedical specialists.  % We conducted a comprehensive manual review of all generated questions, filtering out any unreasonable or erroneous content to ensure the benchmark aligns with the expertise and expectations of real-world biomedical specialists.

\subsection{\acs{qa} Showcases }\label{apxsec:qashowcases}

To better illustrate the full complexity and reasoning depth supported by our approach, we provide the following representative examples drawn from our \ac{qa} dataset.

\textit{Example 1: Multi-step context reasoning}
\begin{lstlisting}[]
Q: When performing the Longitudinal two-photon calcium imaging experiment, after transferring the mouse to an animal cage, what condition should the mouse recover from during the subsequent steps? 
A: anesthesia
\end{lstlisting}
This example requires the model to track the experimental subject's state across sequential procedural steps and infer the relevant recovery condition through cross-sentence coreference resolution, rather than extracting a locally available answer.

\textit{Example 2: Parameter-dependent inference}
\begin{lstlisting}[]
Q: When performing the COVseq experiment, if we use a PCR thermocycler with the lid set at 80 °C, at what temperature should the incubation be performed? 
A: 50 °C
\end{lstlisting}
This \ac{qa} pair reflects a conditional dependency between device configuration and a downstream protocol parameter. The model must distinguish between two co-occurring temperature values and correctly identify which is determined by the given device setting.

\textit{Example 3: Operation-specific reasoning}
\begin{lstlisting}[]
Q: When performing the Fast in vitro protocol for visualization and quantitative high-throughput analysis of sprouting angiogenesis experiment, given that we are imaging beads in large numbers, what unit or entity should the scanning time of about 3 minutes be applied to? 
A: bead
\end{lstlisting}
Answering correctly requires jointly grounding the imaging target, the quantitative scale ("large numbers"), and the time specification — making it a multi-element integration task rather than a simple fact lookup.

\textit{Example 4: Protocol optimization logic}
\begin{lstlisting}[]
Q: When performing the Panel Optimization for High-Dimensional Immunophenotyping experiment, if the overall pattern is incorrect and we need to adjust other laser lines, what should be reduced as part of this adjustment process?
A: respective detector arrays
\end{lstlisting}
This example involves procedural branching logic: the question presupposes a failure condition ("incorrect pattern") and asks the model to reason about the corrective action within a multi-laser optimization workflow, requiring understanding of protocol-level decision structure.

\subsection{Detailed \acs{qa} Evaluation Results}\label{apxsec:detailqa}

\begin{table}[t]
\centering
\resizebox{\linewidth}{!}{%
\begin{tabular}{lcccc}
\toprule
Method & 2 Hop & 3 Hop & 4 Hop & \ac{msr} \\
\midrule
\ac{llm} only          & 18.00 & 5.26  & 0.00  & 15.45 \\
\ac{llm} \acs{lora}    & 12.00 & 0.00  & 0.00  & 9.76  \\
\midrule
BM25              & 56.00 & 36.84 & 25.00 & 52.03 \\
LaBSE             & 59.00 & 31.58 & 25.00 & 53.66 \\
Emb-3-large       & 63.00 & 26.32 & 25.00 & 56.10 \\
Emb-v4            & 63.00 & 21.05 & 0.00  & 54.47 \\
\midrule
GRAG              & 9.00  & 0.00  & 0.00  & 7.32  \\
GRAG \acs{lora}   & 18.00 & 10.53 & 0.00  & 16.26 \\
ToG                & 72.00 & 63.16 & 50.00 & 69.92 \\
\midrule
Ours w/o Sentence        & 67.00 & 57.89 & 50.00 & 65.04 \\
Ours w/o Graph           & 67.00 & 57.89 & 50.00 & 60.16 \\
Ours w/o Planner   & 62.00 & 52.63 & 50.00 & 62.60 \\
Ours w SciERC            & 62.00 & 52.63 & 25.00 & 59.35 \\
Ours w ChemProt    & 68.00 & 52.63 & 25.00 & 64.23 \\
Ours                     & \textbf{76.00} & \textbf{73.68} & \textbf{50.00} & \textbf{74.80} \\
\bottomrule
\end{tabular}
}
\caption{\textbf{Performance comparison across different reasoning depths on the \ac{msr} dataset.}}
\label{tab:reasoning_depth}
\end{table}

\begin{table}[t]
\centering
\resizebox{0.87\linewidth}{!}{%
\begin{tabular}{lcc}
\toprule
& Supervised \ac{ie} & \ac{llm} \ac{ie}  \\
\midrule
\ac{llm} only   & \multicolumn{2}{c}{14.74} \\
\ac{llm} \ac{lora}   & \multicolumn{2}{c}{12.44} \\
\midrule
BM25       & \multicolumn{2}{c}{63.72} \\
LaBSE      & \multicolumn{2}{c}{55.60} \\
Emb-3-large& \multicolumn{2}{c}{61.52} \\
Emb-v4     & \multicolumn{2}{c}{59.14} \\
\midrule
GRAG       & 8.83  & 21.54 \\
GRAG \ac{lora}  & 22.77 & 26.83 \\
ToG              & 69.90 & 65.31 \\
\midrule
Ours w/o Sentence      & 59.84 & 58.25 \\
Ours w/o Graph         & 55.69 & 55.60 \\
Ours w/o Planner & 72.29 & 71.76 \\
Ours w SciERC          & 62.40 & 63.55 \\
Ours w ChemProt        & 64.70 & 62.93 \\
Ours                   & \textbf{72.99} & \textbf{71.76} \\
\bottomrule
\end{tabular}
}
\caption{\textbf{Performance comparison across different \ac{qa} systems.} \textbf{Bold} numbers indicate the best performance among all models.}
\label{tab:ragbenchmark}
\end{table}

\begin{figure*}[t]
\centering
\includegraphics[width=0.95\textwidth]{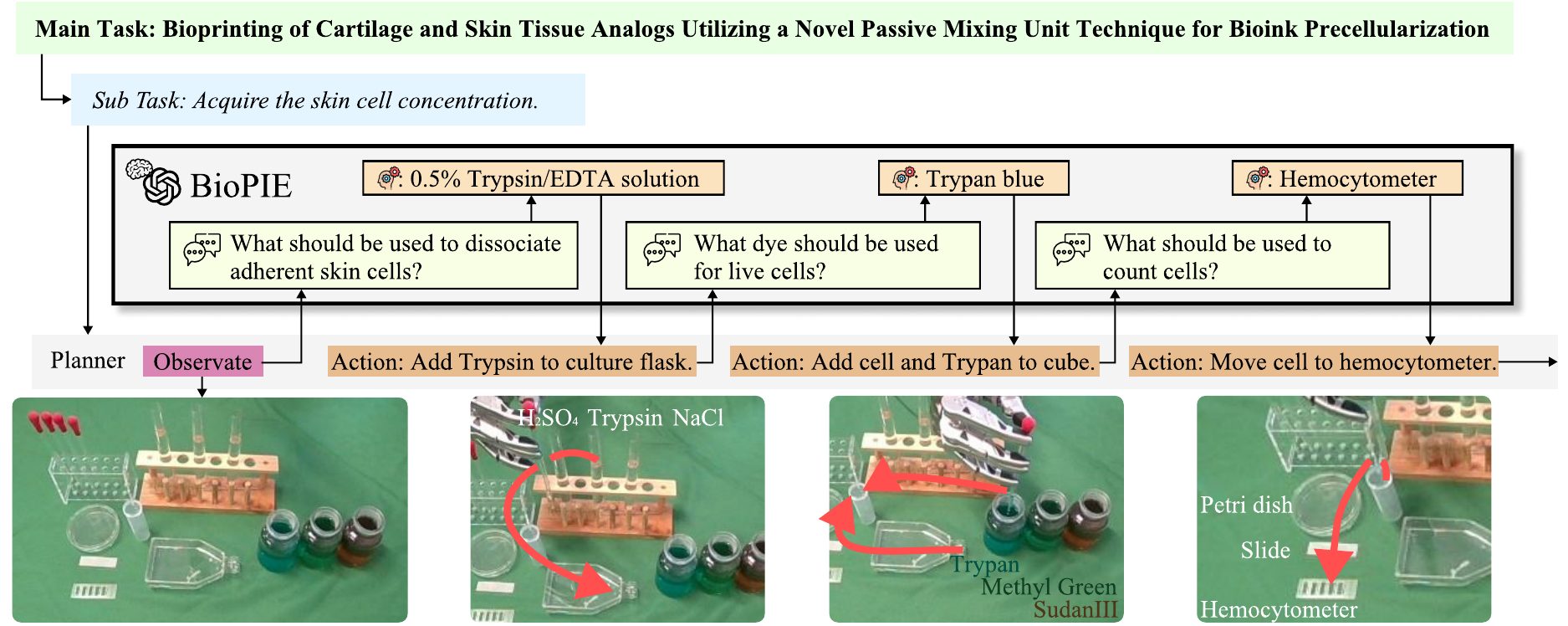}
\caption{\textbf{\ac{biopie} enables knowledge integration in the lab automation.} \ac{biopie} can be used to extract large volume of biomedical protocols into structured knowledge, which can then be used by knowledge systems.}
\label{fig:application}
\end{figure*}

Experiments are conducted on Llama-3-8B. For the \ac{ie} component of our method, we employ the best-performing supervised and \ac{llm}-based extraction approaches under the strict \ac{ood} evaluation setting. 
Specifically, we use HGERE~\citep{yan2023joint} as the supervised \ac{ie} method and Llama-3-8B (Pipeline) as the \ac{llm} \ac{ie} method. 

We have included the performance of the \ac{llm} relative to increasing reasoning depth (see Tab. \ref{tab:reasoning_depth}, \acp{kg} come from supervised \ac{ie}), which demonstrates a downward trend in accuracy with increasing reasoning complexity.

Tab.~\ref{tab:ragbenchmark} reports the overall performance comparison across different \ac{qa} systems. Our method maintains strong performance under both supervised and \ac{llm}-based \ac{ie} settings. Although supervised extraction generally performs slightly better, the performance gap remains small, indicating that the proposed framework is robust to different \ac{ie} strategies.

Although GRAG leverages structured \acp{kg}, its performance remains substantially lower than that of text-based \ac{rag} methods. This can be attributed to the use of average pooling for aggregating node representations, which may limit the model's ability to capture fine-grained and localized subgraph semantics. Consequently, the retrieved subgraphs often provide insufficient descriptive information, leading to consistently lower retrieval hit rates (see Fig. \ref{fig:qacompare}\textbf{(B)}). In contrast, biomedical experimental \ac{qa} typically involves a large number of domain-specific terms, in which text-based retrievers naturally achieve a higher recall and more reliable evidence retrieval.

\subsection{Application Showcase}\label{apxsec:qaapplication}

In this subsection, we demonstrate a practical application of \ac{biopie} in the field of lab automation. As illustrated in Fig. \ref{fig:application}, \ac{biopie} serves as a knowledge integration engine that bridges the gap between high-level biomedical protocols and robotic execution. By extracting structured knowledge from vast volumes of scientific literature, the system can provide precise answers to critical procedural questions—such as identifying the correct reagents for cell dissociation or the appropriate dyes for viability assays. Such structured understanding enables an automated planner to orchestrate a sequence of precise laboratory actions, ranging from trypsinization to cell counting using a hemocytometer, thereby facilitating the complex bioprinting of cartilage and skin tissue analogs with minimal human intervention.

\end{document}